\documentclass[sigconf,nonacm]{acmart}
\usepackage{multirow}
\usepackage{booktabs}
\usepackage{float}
\usepackage{amsmath}
\usepackage{makecell}
\usepackage{enumitem}
\usepackage{graphicx}
\usepackage{lipsum}
\usepackage[export]{adjustbox}

\AtBeginDocument{%
  \providecommand\BibTeX{{%
    \normalfont B\kern-0.5em{\scshape i\kern-0.25em b}\kern-0.8em\TeX}}}

\makeatletter
\def\@copyrightspace{\relax}
\makeatother
\begin{document}

% \title{RJUA-MedDQA: A benchmark for Visual Question Answering and Contextual Diagnostic Reasoning}
\title{RJUA-MedDQA: A Multimodal Benchmark for Medical Document Question Answering and Clinical Reasoning
}

\author{Congyun Jin} 
\authornote{Equal Contribution}
\author{Yujiao Li}
\author{Yingbo Wang}
\author{Yabo Jia}
\affiliation{%
  \institution{Ant Group}
  \city{Shanghai}
  \country{China}
}

\author{Ming Zhang}
\authornotemark[1]
 \author{Weixiao Ma}
\author{Chenfei Chi}
\author{Xiangguo Lv}
\affiliation{%
  \institution{Renji Hospital}
  \city{Shanghai}
  \country{China}
  \footnote[4]
  }

\author{Fangzhou Li}
\author{Wei Xue} 
\author{Yiran Huang}
\authornote{Yiran Huang is the corresponding author (huangyiran@renji.com)}
\affiliation{%
  \institution{Renji Hospital}
  \city{Shanghai}
  \country{China}
  \authornote{Department of Urology, Shanghai Jiao
Tong University School of Medicine Affiliated Renji Hospital}\footnote[4]
  }

\author{Tao Sun}
\author{Haowen Wang}
\author{Yuliang Du} 
\author{Cong Fan}
\authornote{Cong Fan is the corresponding author (fancong.fan@antgroup.com)}
\author{Jinjie Gu}
\affiliation{%
  \institution{Ant Group}
  \city{Shanghai}
  \country{China}
}

\renewcommand{\shortauthors}{Jin, et al.}


\begin{abstract}
% With the recent progress in  
% Medical report agent, accurate document layout 

% disease diagnosis and disease state diagnois

% We introduce 
% reveal where they struggle most.

% Layout-aware pre-trained models has achieved significant
% progress on document image question answering

%  In this paper, we present
% DocLayNet, a new, publicly available, document-layout annotation
% dataset in COCO format. It contains 80863 manually annotated
% pages from diverse data sources to represent a wide variability in
% layouts. 
Recent advancements in Large Language Models (LLMs) and Large Multi-modal Models (LMMs) have shown potential in various medical applications, such as Intelligent Medical Diagnosis. Although impressive results have been achieved, we find that existing benchmarks do not reflect the complexity of real medical reports and specialized in-depth reasoning capabilities. In this work, we establish a comprehensive benchmark in the field of medical specialization and introduced RJUA-MedDQA, which contains 2000 real-world Chinese medical report images poses several challenges: comprehensively interpreting imgage content across a wide variety of challenging layouts, possessing the numerical reasoning ability to identify abnormal indicators and demonstrating robust clinical reasoning ability to provide the statement of disease diagnosis, status and advice based on a collection of medical contexts. We carefully design the data generation pipeline and proposed the Efficient Structural Restoration Annotation (ESRA) Method, aimed at restoring textual and tabular content in medical report images. This method substantially enhances annotation efficiency, doubling the productivity of each annotator, and yields a 26.8\% improvement in accuracy. We conduct extensive evaluations, including few-shot assessments of 5 LMMs which are capable of solving Chinese medical QA tasks. To further investigate the limitations and potential of current LMMs, we conduct comparative experiments on a set of strong LLMs by using image-text generated by ESRA method. We report the performance of baselines and offer several observations: (1) The overall performance of existing LMMs is still limited; however LMMs more robust to low-quality and diverse-structured images compared to LLMs. (3) Reasoning across context and image content present significant challenges. We hope this benchmark helps the community make progress on these challenging tasks in multi-modal medical document understanding and facilitate its application in healthcare. 

% Our dataset will be publicly available for noncommercial use at 
\end{abstract}





\maketitle

\section{Introduction}

The breakthroughs in large language models (LLMs) bring generalist AI models that can solve a wide range of medical challenges such as Automatic Diagnosis Agent \cite{liu2022my,he2023survey}. Large multi-modal models (LMMs) aim to achieve even stronger general intelligence via extending LLMs with multi-modal inputs such as Medical Imaging Analysis \cite{Moon_2022, xu2023multimodal}. Consequently, advances in Vision-Language multi-modal research can be beneficial in assisting in clinical decision-making, improving patient engagement, and helping relieve the burden of the healthcare system and improve medical professionals’ efficiency \cite{LIN2023102611}.

Although these endeavors have led to remarkable strides forward, medical report understanding has received little attention so far, though it offers great potential for practical use since medical report images emerge as a powerful tool, offering a clear, concise, and accessible way to communicate vital clinical information including patient's basic information, examination details, abnormal test indicators and disease diagnosis. 
% To train models for understanding a medical report, the datasets are mainly collected from general PDF document Visual Question Answering (VQA) dataset.
Generally, there are several available datasets with document-based VQA or reasoning-based VQA. Despite these important advances, limitations exist: (1) In currently available document-based datasets \cite{mathew2021docvqa, ding2023pdfvqa, Pfitzmann_2022}, most do not require reasoning ability. A few requires discrete reasoning based on image content such as TAT-DQA \cite{Zhu_2022} There is a lack of contextual reasoning dataset. (2) In currently available document VQAs and contextual-reasoning dataset, there are a lack of real-world medical visual document and specialized medical data relevant to diagnosis, status and advice of diseases. Meanwhile, the current LMM models still have trouble understanding information within a real-world image which usually diverse layout, skewed and blurry. Moreover, compared with humans, large LMMs are still not capable of performing numerical or contextual reasoning with successively logical connections within entire image.
% The existing reasoning VQA limits the scale of document understanding to a single independent document page.  
% It is a more natural demand to holistically understand the full image and capture the connections of textual contents and their casual relationships across the entire page. 
Based on these observations, we introduce a novel benchmark, RJUA-MedDQA, which are sampled from Shanghai Renji Hospital within Urology Department Database. The raw medical reports were subsequently processed and anonymized to create synthetic multi-modal clinical data for virtual patients, ensuring that no medical confidentiality or patient privacy is compromised. RJUA-MedDQA is designed to reflect the challenges encountered in practice. We hope that this benchmark helps bridge the gap between academic research and practical scenarios to facilitate future study on multi-model medical document understanding.

\begin{figure}
    \centering
    \includegraphics[width=1\linewidth]{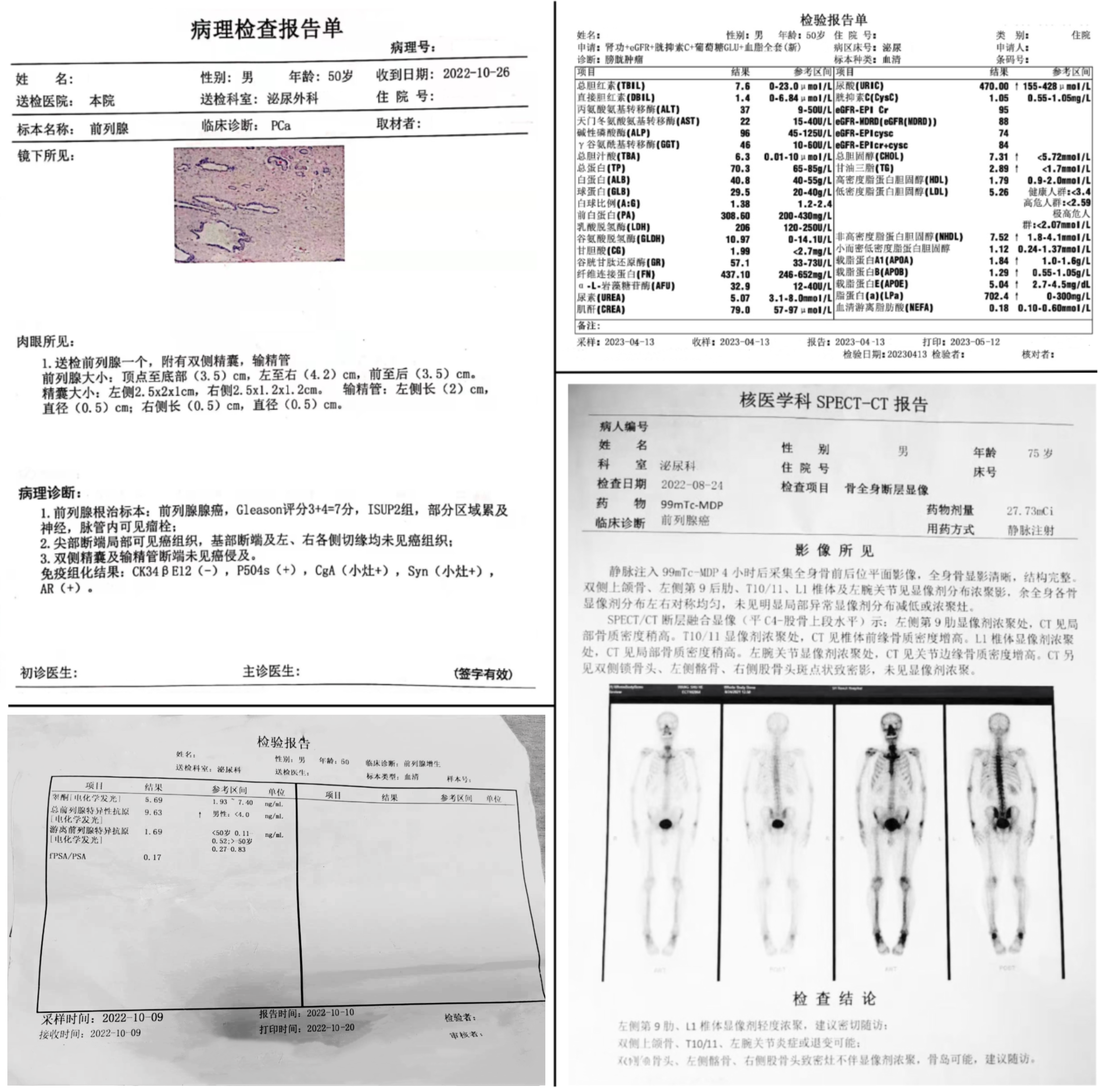}
    \caption{Complex Page Layouts in Various Medical Report Categories: Four Illustrative Examples}
    \label{fig:layout}
\end{figure}

The dataset is collaboratively constructed by a team of urological experts from Shanghai Renji Hospital, comprising over 2,000 images of real-world medical report featuring a range of complex layouts that include both textual and tabular information, shown in Figure ~\ref{fig:layout}.
RJUA-MedDQA is a high-level document understanding task wherein given a visually-rich medical report and a relevant question in natural language, and may attached with a set of facts or evidences extracted from official medical guideline or clinical experience. The model is required to give the correct answer to the question based on the image. 

Our main contributions are summarized as follows:
\begin{itemize}[leftmargin=4.5mm]
    \item \textbf{The Largest Medical Report Benchmark in Chinese:} RJUA-MedDQA is a comprehensive benchmark for visually-rich medical report understanding in Chinese, with a focus on Urology. To the best of our knowledge, this is also the largest real-world medical report VQA dataset. We decide to release a portion of the dataset with high-quality OCR results and annotations. Additionally, we define three tasks corresponding to different application scenarios. The primary goal of RJUA-MedDQA is to enhance LMMs to understand medical reports, enabling precise interpretation of content across varied diverse layouts and strong logical reasoning given a list of medical knowledge.
    \item \textbf{Large Layout Variability:} The dataset comprises a variety of image types, including photographs, scanned PDFs, and screenshots that are not only characterized by their diverse and complex layouts from numerous public sources, but also by varying image quality. The photographed and screenshot images, in particular, may display signs of diminished quality due to factors, like rotation, skewing, blurriness of text, or missing information, highlighting challenges encountered in real-world scenarios.
    \item \textbf{Efficient Structural Restoration Annotation (ESRA) Method:} In contrast to conventional layout annotation methods, we propose an efficient structure-aware labeling method to reconstruct both textual and tabular content in medical reports. It significantly reduces the human labeling errors and improves the efficiency of annotation process. Statistically, this method has elevated the accuracy rate from 70\% to 96.8\%.
    \item \textbf{Synonym-aware Automatic QA Generator:} Given the ESRA method, we further integrated an synonym-aware automatic QA generator with an extensive range of templates, which are capable of handling tasks ranging from simple fact retrieval to more complex inference-based questions. Additionally, these templates are highly flexible to be customized based on given annotations, providing the capability to fine-tune or evaluate models on specialized research inquiries.
    \item \textbf{Clinical Expert Annotation:}  The dataset is also precisely annotated by urology specialists for contextual reasoning task. This dataset also provides an fact base, which includes logical chain on the diagnosis of disease and disease stage, and treatment advises, mainly extracted from clinical experience and official Urological Disease Diagnosis and Treatment Guideline \cite{huang}. This fact base attempts to mitigate the gap between urological disease diagnosis and research communities. 
\end{itemize}

% A practical medical report question is usually a combination of consultation, diagnosis and treatment.

% to accurately describe the detailed and structured process during a badminton match
% A practical medical report question is usually a combination of consultation, diagnosis and treatment. Thus, it is significant to expand the current scale of page-level document understanding to the full document-level. To the best of our knowledge, no previous study considers all three medical services simultaneously.

\section{Related Work}

\begin{table*}[]
\caption{Comparison of VQAs. Answer types can be broken down into single-span(SS), multi-span(MS), and non-span(NS). “S/P” denotes the “Scanned/Photographed” modality of images.}
\label{tab:dataset}
\small
\begin{tabular}{@{}ccccccccc@{}}
\toprule
\textbf{Dataset} & \multicolumn{1}{c}{\textbf{Domain}} & \multicolumn{1}{c}{\textbf{\#Images}} & \multicolumn{1}{c}{\textbf{\begin{tabular}[c]{@{}c@{}}Image\\ Type\end{tabular}}} & \multicolumn{1}{c}{\textbf{\begin{tabular}[c]{@{}c@{}}Images Modal\\Type\end{tabular}}} & \multicolumn{1}{c}{\textbf{\begin{tabular}[c]{@{}c@{}}w/o Context\\ Reasoning\end{tabular}}} & \multicolumn{1}{c}{\textbf{\begin{tabular}[c]{@{}c@{}}w/ Context\\ Reasoning\end{tabular}}} & \multicolumn{1}{c}{\textbf{\#QAs}} & \multicolumn{1}{c}{\textbf{\begin{tabular}[c]{@{}c@{}}Answer\\ Type\end{tabular}}}  \\ \midrule
DocVQA & General & 12k & S & \multicolumn{1}{c}{Document} &  &  & 50k & SS   \\
TAT-DQA & Finance & 2k & S &Document&\checkmark  &  & 1.4k & SS,MS,NS   \\
InfographicVQA & General & 5k & S &Document&\checkmark  &  & 30k & SS,MS,NS   \\
SlideVQA & General & 52k & S &Document&\checkmark  &  & 14.5k & SS,MS,NS  \\
MultiModalQA& General & 57k& S &Document  & \checkmark  & & 30k& SS,MS,NS  \\
ScienceQA & General & 6.5k & S & Non-document&  &\checkmark  & 6.5k & NS   \\
A-OKVQA & General & 2.9k & S & Non-document &  & \checkmark & 2.9k & NS   \\
SLAKE & Medical & 642 & S & Non-document &\checkmark  &  & 14k & NS   \\
\textbf{RJUA-MedDQA}& Medical & 2k & S/P & Document&\checkmark  & \checkmark & 72k  & SS,MS,NS   \\ \bottomrule
\end{tabular}
\end{table*}

In this section, we briefly review previous research in the datasets for Document VQA (DQA), reasoning over VQA and medical VQA, with special attention to those works that are most related to ours. The properties of these datasets and the comparison with ours are shown in Table ~\ref{tab:dataset}.

\textbf{Document VQA}
Document VQA is a high-level document understanding task wherein a model is required to answer a question in natural language given a visually-rich document. Some useful datasets have been published. DocVQA is constructed using various types of industry documents for extractive question answering, where answers can always be extracted from the text in the document \cite{mathew2021docvqa}. VisualMRC dataset is built for abstractive question answering, where answers cannot be directly extracted from the text in the document \cite{tanaka2021visualmrc}. Meanwhile, more datasets are moving beyond extractive question answering and are increasingly focusing on the reasoning capabilities over given image, including InfographicVQA \cite{mathew2021infographicvqa}, SlideVQA \cite{tanaka2023slidevqa}, TAT-DQA \cite{Zhu_2022} and MultiModalQA \cite{talmor2021multimodalqa}. All of these datasets require single-hop or multi-hop reasoning capabilities. Among them, InfographicVQA focuses on infographic instead of documents; SlideVQA focuses slide decks instead of documents; TAT-DQA dataset includes real-world high-quality business documents and MultiModalQA requires joint reasoning over textual and visual information in Wikipedia, while our RJUA-MedDQA dataset includes real-world medical documents of varying quality.

\textbf{Contextual Reasoning VQA} The previous task assumes that the datasets have a relevant image, containing all the facts required to answer. In Contextual Reasoning VQA task, the image may not all the facts to answer a question. Instead, a rationale or a chunk of paragraph is given to provide the knowledge supplements each question: Context-VQA \cite{naik2023contextvqa}, ScienceQA \cite{lu2022learn} and A-OKVQA \cite{schwenk2022aokvqa}. In Context-VQA, context refers to the type of website from which an image is sourced, which differs from our use of the term to describe additional data that aids in answering visual questions. Meanwhile, the context provided in ScienceQA is artificially-generated. However, all existing contextual Reasoning VQA contains visual or simple textual information but no layout information.

\textbf{Medical VQA} There are several public-available Medical VQA datasets up to date.
VQA-MED-2018 \cite{Hasan2018OverviewOI}, VQA-RAD \cite{VQARAD}, VQAMED-2019 \cite{ImageCLEFVQA-Med2019}, RadVisDial \cite{kovaleva-etal-2020-towards}, PathVQA \cite{he2020pathvqa}, VQA-MED-2020 \cite{ImageCLEFVQA-Med2020}, SLAKE \cite{liu2021slake}, and VQA-MED-2021 \cite{ImageCLEF-VQA-Med2021}. The selected datasets are diverse in image modality and question categories. The imaging modality of those datasets covers chest X-ray, CT, MRI, and pathology. The questions include close-end questions and open-end questions on a variety of topics \cite{Lin_2023}.  SLAKE is the most similar to us, as it is a comprehensive dataset with both semantic labels and a structural medical knowledge base. 
The medical knowledge base is presented as a knowledge graph, instead of being specific to each individual question. Furthermore, all these medical datasets comprise exclusively of visual image.

Therefore, we introduce RJUA-MedDQA for document-content understanding and contextual diagnosis reasoning. Our objective is to advance research in multi-modal document VQA within the academic community and to support the medical community in developing applications that improve their clinical reasoning abilities.

\section{The Dataset}
\begin{figure}
    \centering
    \includegraphics[width=1\linewidth]{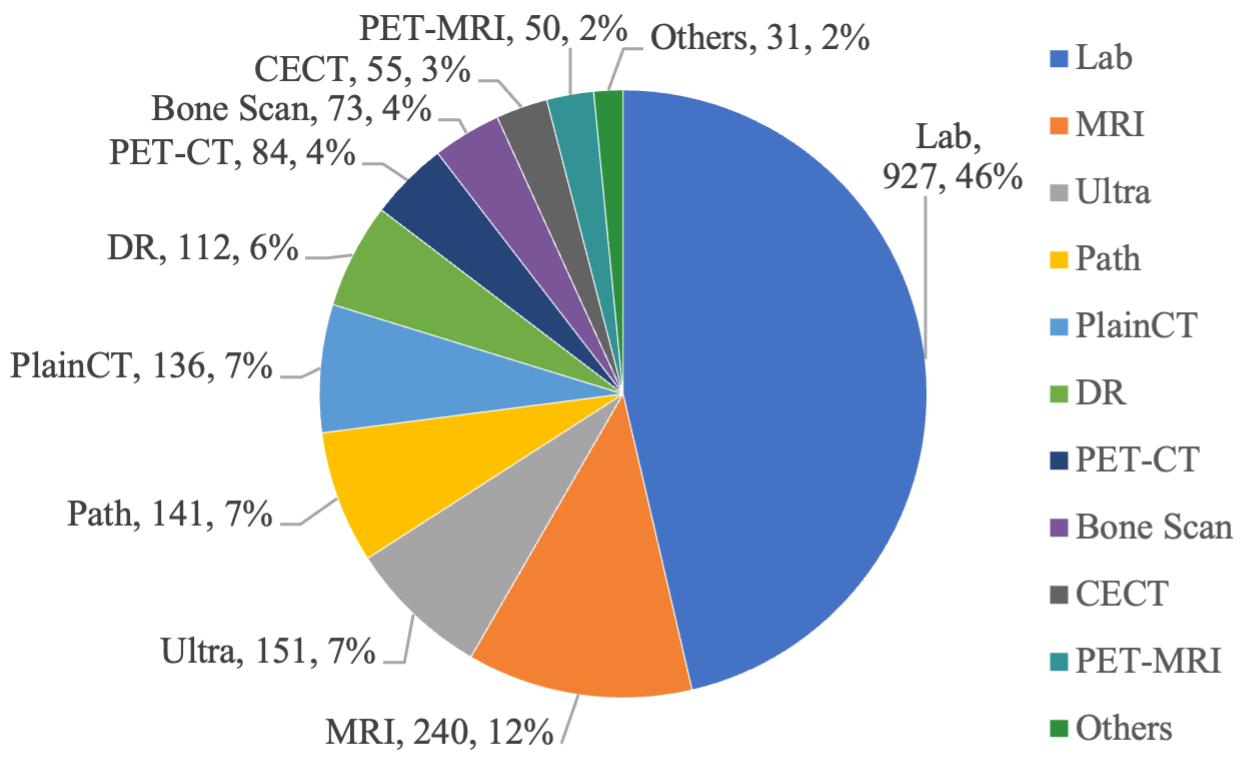}
    \caption{Statistics by report types}
    \label{fig:2}
\end{figure}
\subsection{Overview}
The RJUA-MedDQA dataset contains a total of 2000 images, of which 402 are screenshot, 619 are scanned-PDF, and the remaining 979 are photos taken by patients during real-world medical consultations. Reports in screenshot and scanned-PDF format ensure the integrity and clarity of information; on the other hand, reports captured in photographs may exhibit some degree of quality degradation caused by issues such as rotated or skewed angles, blurred text, or incomplete information, which reveals real-world problems. Medical reports can be grouped into two main categories, namely Laboratory Report and Clinical Report. Figure ~\ref{fig:2} shows the overall frequency and distribution of different report types.

\textbf{Laboratory Reports:} mainly include complete blood count (CBC), liver function tests (LFTs), urinalysis, and urological disease screening tests such as Prostate-Specific Antigen (PSA) Testing, which comprising a well-structured table and some relevant texts. Tables demonstrate large variability, ranging from single-column to multiple-column formats, with or without lines. Primarily, they convey laboratory results and its corresponding reference range.

\textbf{Diagnostic Reports:} mainly include
MRI, Ultrasound, Pathology, Digital Radiography(DR), PlainCT, Contrast-enhanced CT (CECT), as well as other diagnostic modalities like PET-CT, PET-MRI, Endoscopy Report and Renogram. 

Overall, the research team has compiled statistics for the year 2023, from Renji Hospital's Department of Urology in Shanghai, encompassing initial diagnoses of urological diseases among 319,401 patients. They conducted 934,675 major laboratory tests, and a combined total of 401,615 examinations and pathological assessments. The dataset covers the top 50 most frequent laboratory tests to 95\% (48/50) and the top 20 most frequent examinations and pathological procedures to 98\% (18/20), respectively. Furthermore, the dataset covers over 334 different diseases and diagnoses, with detailed statistics provided in Appendix ~\ref{diease:statistics}. Characteristically, these reports vary widely in their formatting, often including an extensive range of medical terminology and related synonyms, thereby achieving a high level of variability and the capacity for generalization.
% Collectively, these categories form a set of 11 distinct types of reports, which together constitute approximately 95\% of all report types generated in a hospital. 

%also provide additional annotations in bounding box and

\begin{figure*}[t]
    \centering
    \includegraphics[width=1\linewidth]{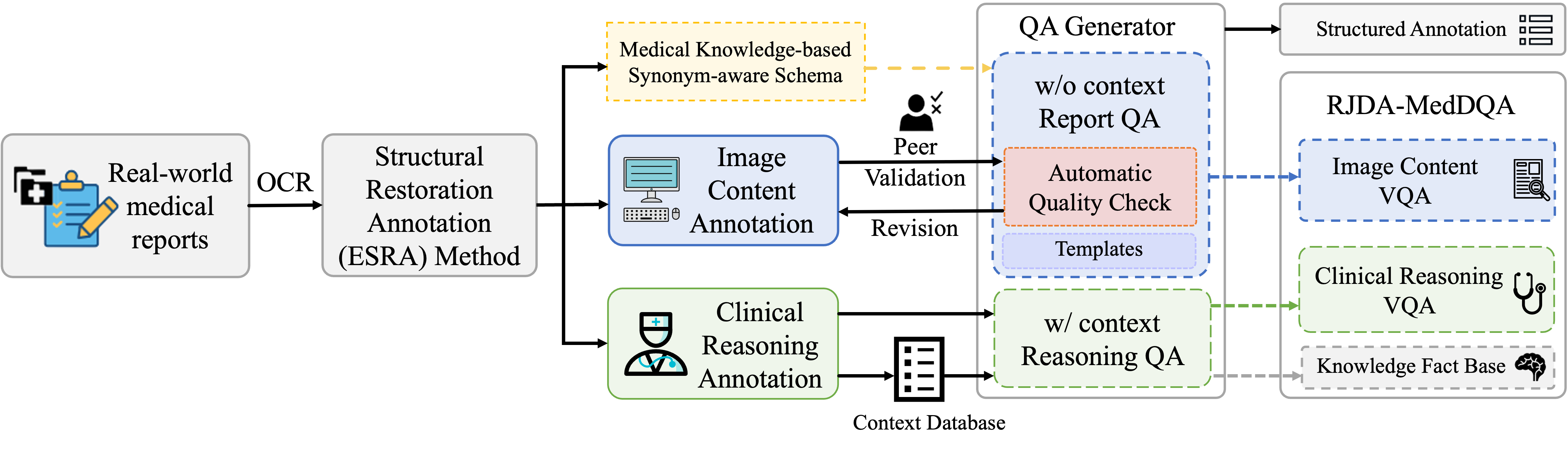}
    \caption{Data Generation Pipeline}
    \label{fig:3}
\end{figure*}

\subsection{Task Overview and Definition}
We introduce RJUA-MedDQA dataset for the medical report understanding question-answering problem requiring models to possess the capability to interpret textual and tabular content within images, as well as reasoning capacity given a chunk of context. Consider a medical report $D$, which contains text content and possibly including a table, we propose two main tasks: (1) Non-Contextual QA in Report Comprehension; (2) Contextual QA in Clinical Reasoning. 

\subsubsection{Task 1: Non-Contextual QA in Report Comprehension} 
Given a question $Q$, the model $\mathcal{F}$ is required to predict the answer $a$ according to the report $D$. Formally, the task is formulated as
\begin{equation}
    \mathcal{F}(D,Q)=a
\end{equation}
In w/o context VQA, the answer may be either extracted from the given image, or generated by performing mathematical reasoning. To solve the former task, a LMM model needs to possess strong ability to interpret content and tabular layout in order to derive the final answer.  In the second scenario, there is a demand for discrete reasoning capability across the entire visually-rich report.

\subsubsection{Task 2: Contextual QA in Clinical Reasoning} 
Given a question $Q$, and a piece of context $C$ that includes the gold facts necessary for answering $Q$ and distracting facts, which is formulated as
\begin{equation}
    \mathcal{F}(D,C,Q)=a
\end{equation}
In w/ context VQA, the answer can be derived based on patient's basic information such as age, examination description and the evidence in the provided context. Questions include (1) Disease diagnosis based on abnormal indicators in laboratory report or examination findings in non-laboratory test report; (2) The current severity of the disease, such as staging of prostate cancer; (3) Definitive treatment advice. In this process, the capability of logical reasoning over the visual-language modality is much demanded.

\section{Data Generation Pipeline}
The Data Generation Pipeline was carried out in three phases.
In phase one, we firstly identify and prepare the data sources, and then apply the Efficient Structural Restoration Annotation (ESRA) Method to structurally recovered textual and tabular content based on OCR results. In phrase two, we carefully design the annotation guideline in order to obtain high efficiency and maximum consistency, and then upload each image along with its corresponding image-text to an online interface. The annotation process is mainly divided into two parts: 1) Image Content Annotation; 2) Clinical Reasoning Annotation. Both of them includes a efficient and detailed guideline with the format of the data, the specific labeling task and the examples of various types of labels. We performed continuous quality controls. In phrase three, reviewed annotations are refined through an automated generator to produce the final dataset. A illustration is shown in Figure ~\ref{fig:3}.  

% The annotation process was carried out in two phases. In phase one, we identified and prepared the data sources for annotation. In phase two, we determined how annotations should be designed in order to obtain high efficiency and maximum consistency.  
% The annotation process is mainly divided
% into two parts: 1) Image Content Annotation; 2) Clinical Reasoning Annotation, and both of them includes a efficient and detailed guideline with the format of the data, the specific labeling task and the examples of various types of labels.

% In phase three, we performed continuous quality controls.

\subsection{Efficient Structural Restoration Annotation}
\label{ESRA}
\begin{figure*}[htbp]
\centering\includegraphics[width=\textwidth]{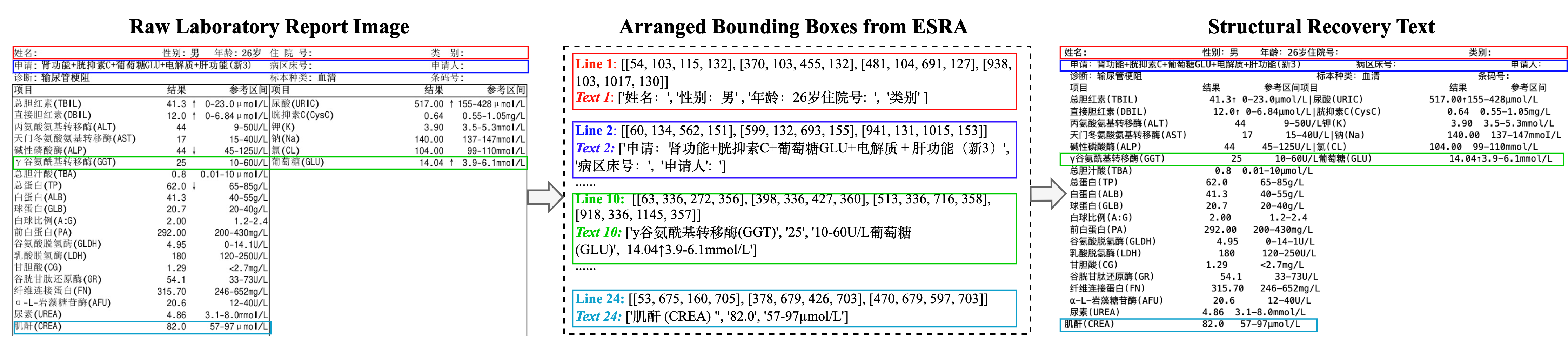}
    \caption{A Demonstration of ESRA method}
    \label{fig:4}
\end{figure*}
Given OCR results, the information of coordinates are converted to appropriate spaces and line breaks which connect all chunks of texts, resulting in structural text that is similar to the original medical report image(see Figure ~\ref{fig:4}). 

Mathematically, given a medical report image $I$, an OCR tool is applied to extract textual content. The resulting OCR results consist of extracted text segments and their corresponding bounding boxes (bbox) are denoted as $T=\{t_1,t_2,...,t_n\}$ and $B=\{b_1,b_2,...,b_n\}$, where $n$ represents the number of text segments recognized.

\begin{itemize}
\item[Step 1] Based on the bbox $B=\{b_1, b_2, ..., b_n\}$, partition the boxes that belong to the same text line. Specifically, the coordinates within $B$ are sorted orderly from left to right, and then top to bottom, resulting new text segments $T^*=\{t_1^*,t_2^*,...,t_n^*\}$ and new bbox $B^*=\{b_1^*,b_2^*,...,b_n^*\}$, where $b_i^*=[x_{i,0},x_{i,1},y_{i,0},y_{i,1}]$
\item[Step 2] The bbox in $B$ are traversed and converted into lines denoted as $line_i$ by line flag.
\begin{equation}
\begin{aligned}
    flag=bool[&(y_{i+1,0}+\varepsilon_1 < \frac{y_{i,1}-y_{i,0}}{2}) \mid \\
    &(y_{i,0}+\varepsilon_2 < \frac{y_{i+1,1}-y_{i+1,0}}{2}<y_{i,1}-\varepsilon_2)]
\end{aligned}
\end{equation}

where $\varepsilon_1=r^* (y_{i,1}-y_{i,0})$ and 
$\varepsilon_2=r^* (y_{i+1,1}-y_{i+1,0})$. $r^*$ is a discount coefficient ranging from 0 to 1. By setting this hyper-parameter $r^*$, the adhesion of text segments between lines can be mitigated, thereby increasing readability.

If $flag=True$, bbox $b_i$ and $b_{i+1}$ are on the same line. 

If $flag=False$, a new line will be added, resulting bbox 
\begin{equation}
\begin{aligned}
    B^*=\{line_1,line_2,... ,line_n\}
\end{aligned}
\end{equation}
where $line_i=\{b^*_{line_{i1}},b^*_{line_{i2}},...,b^*_{line_{in}}\}$. 
\item[Step 3] Define a set $\mathcal{H}$ representing the heights of all bbox of $line_i$. Apply k-means clustering to $\mathcal{H}$ on generate clusters. Within each cluster $c_i$, determine the character count $N_i$ of bbox. Identify the cluster $c_{max}$ that has the maximum character count. The total width of bbox within $c_{max}$ is denoted as $w^*_{total}$ and total characters in $c_{max}$ is denoted as $N^*_{total}$. The average character width $c^*$ can be calculated as:
\begin{equation}
    c^*= \frac{w^*_{total}}{N^*_{total}}
\end{equation}

\item[Step 4] Join text segments in the same line by spaces. Given two adjacent text segments $b^*_{line_{ij}}$,$b^*_{line_{ik}}$

\begin{equation}
    \text{number\_of\_spaces} = \max(\frac{h_{i,j,k}}{c^*l},1)
\end{equation}

where $h_{i,j,k}$ is the horizontal distance between the bbox $b^*_{line_{ij}}$,$b^*_{line_{ik}}$.
Additionally, $l$ is an expansion coefficient from 0 to 1. By setting the hyper-parameter $l$, we control the coefficient that determines the number of spaces per line in the overall layout output, thereby increasing integrity.
\end{itemize}

A demonstration of the impact of different $r$ on the output structure of ESRA method and detailed hyper-parameters setting can be found in Appendix ~\ref{hyperparameter}.

\subsection{Annotation Process}
All images are uploading to a online platform which provides a visual annotation interface and allows for dataset inspection and analysis. 
% Raw medical reports are pre-labeled with report type tag based on OCR results mentioned in section ~\ref{ESRA}. Reports are then grouped by categories, facilitating physicians in marking disease diagnoses according to report details since the medical terminology of the same report type tends to exhibit similarities during clinical reasoning annotation.

\subsubsection{Image Content Annotation}
A group of 10 dedicated annotators with basic medical knowledge were responsible for this part to ensure the accuracy and professionalism of data labeling. We review the collected reports, identify the most common structural features they exhibit, and choose two most representative aspects: 
\begin{itemize}[leftmargin=4.5mm]
    \item Key-Value Pairs: Annotators are required to identify and extract key information pairs from images, such as age, examination date, pre-examination clinical diagnosis findings, etc., and label them as structured fields.
    \item Quadruplets <item, result, range, is\_abnormal> Annotators are required to restoring the structure of tables, including the name of item and the corresponding result and reference interval, and mark abnormality. Each indicator can be labeled as normal, abnormal, or undetermined—the latter if no reference range is given. For abnormal results, annotators must also specify if the value is high or low.
\end{itemize}
All contents in the report image are annotated either Key:value pairs or Quadruplets. During the labeling process, annotators need to appropriately fill incomplete information such as the missing keys and partially missing item names to increase the generalization ability of the dataset.
\begin{table*}[]
\caption{Examples of QA generators for a variety of task and difficulties (\textit{Synonyms}) \texttt{Annotations}}
\label{tab:qa_generator_example}
\small
\begin{tabular}{ccll}
\toprule
\textbf{Task} & \textbf{Sub-task} & \multicolumn{1}{c}{\textbf{Question Template}} & \multicolumn{1}{c}{\textbf{Answer Template}} \\ \Xhline{1.5\arrayrulewidth}
\multirow{2}{*}{Entity} & Single & What is  \texttt{key} \textit{(impression)} ? & \texttt{key} is \texttt{value}. \\
 & Multi & What are \texttt{key1} \texttt{(age)} and \texttt{key2} \textit{(date)}? & \texttt{value1}; \texttt{value2} \\ \hline
\multirow{5}{*}{Table} & Single Cell & What is the \textit{(laboratory result)} of \texttt{item1} & The  \textit{(laboratory result)} of \texttt{item1} is \texttt{result1}. \\
 & Single Row & \begin{tabular}[c]{@{}l@{}}What is the \textit{(result)} and\\  \textit{(reference range)} of \texttt{item1} ?\end{tabular} & \begin{tabular}[c]{@{}l@{}}The \textit{(result)} of \texttt{item1} is \texttt{result1},\\ and the \textit{(reference range)} is \texttt{range1}.\end{tabular} \\
 & Multi-Row & \begin{tabular}[c]{@{}l@{}}What are the \textit{(results)} and \textit{(standard interval)} \\ of \texttt{item1} and \texttt{item2} correspondingly?\end{tabular} & \begin{tabular}[c]{@{}l@{}}\texttt{item1}, \texttt{result1}, \texttt{range1};\\ \texttt{item2}, \texttt{result2}, \texttt{range}2;\end{tabular} \\ \hline
\multirow{2}{*}{TableNR} & Comparison & Is item1 within \textit{(normal range)}? & \begin{tabular}[c]{@{}l@{}}The \textit{(result)} of \texttt{item1} is \texttt{result1} and \\ the \textit{(normal range)} is \texttt{range1}, hence  \texttt{is\_abnormal1}\end{tabular} \\
 & Multi & Is there any \textit{(abnormal indicators)} in this report? & \texttt{is\_abnormal1}, \texttt{is\_abnormal3}, \texttt{is\_abnormal9}. \\ \hline
Customized & Summarization & \begin{tabular}[c]{@{}l@{}}What key elements should be \\ noticed in this medical report?\end{tabular} & \begin{tabular}[c]{@{}l@{}}There are \texttt{len({[}items{]})} in this report. \\ \texttt{len(is\_abnormal==1)} are not in standard reference, which are ...\end{tabular} \\ \bottomrule
\end{tabular}
\end{table*}

Overall, ESRA approach greatly diminishes the possibility of human input errors and enhances the efficiency of annotation. It avoids the use of traditional typing by annotators, significantly reducing the error rate that may occur during the typing process and saving huge amount of time for typing obscure medical terminology and descriptions. Additionally, annotators also refine OCR results to further improve accuracy. Consequently, the work efficiency of the standard annotators can be increased from 5 images per hour to 10 images per hour. The accuracy rate has improved from 70\% to 96.8\% by manually inspecting the annotation quality of 100 images.

\subsubsection{Quality Check}
The production phase took around three weeks to complete. Subsequently, a peer-validation was carried out to rectify any errors in the annotations. We further conducted continuous quality checks during data QA generation process by designing an automated quality check tool based on this annotation framework. It includes features to identify issues such as missing keys or values, discrepancies in the number of table items, and the potential for error in the assessment of abnormal indicators. Problematic samples are returned for re-verification and manual annotation correction.

\subsubsection{Clinical Reasoning Annotation}
\label{sec:medical_reasoning_annotation}
A small team of 5 urological experts are assembled for Clinical Reasoning Annotation. It is more time-consuming even with the assistance of ESRA, because it demands the specialized knowledge and clinical experience. To accelerate this process and streamline the subsequent generation of the QA dataset, we firstly established a context database. Meanwhile, in order to mirror the real-world diagnostic process of doctors reviewing reports, context are categorized into: 1) Examination-Disease 2) Examination-Status; 3) Disease-Status; 4) Disease-Advice; 5) Examination; 6) Disease-Examination 7) Disease-Treatment. 

In this way, physicians can add relevant context directly to the database while annotating, by indexing each context a descriptive title. They can quickly reference and tag recurring content, significantly boosting annotation efficiency and consistency. 

Consequently, the labeling of each report involves assigning context indexes for: 1) Diagnosis; 2) Staging or status; and 3) Advice, treatment or supplementary examinations. All labeling statements are unambiguous and unbiased, and all suspicious findings are excluded. A report may contain none, one or more diagnoses, which may be explicitly displayed in report, or could be inferred from the report content and corresponding context. Statistics, analysis and detailed examples of annotated contexts are shown in Appendix ~\ref{context} (see Table~\ref{contextexample}).
\begin{table*}[h]
\caption{Performance of Baseline Models on RJUA-MedDQA. Best results are marked in bold}
\label{table:overall}
\small
\begin{tabular}{cc|ccccc|cc}
\toprule
\multicolumn{2}{c|}{} & \multicolumn{5}{c|}{\textbf{Image   Content}} & \multicolumn{2}{c}{\textbf{Clinical   Reasoning}} \\ \cline{3-9} 
\multicolumn{2}{c|}{} & \multicolumn{2}{c}{Entity} & \multicolumn{2}{c}{Table} & TableNR & MC & SA \\ \cline{3-9} 
\multicolumn{2}{c|}{\multirow{-3}{*}{\textbf{Method}}} & RougeL & Acc & RougeL & Acc & Acc & Acc & RougeL \\ 
\Xhline{1.5\arrayrulewidth}
\multicolumn{1}{c|}{} & LLaVA-v1.5-7B & 0.08 & 0.04 & 0.08 & 0.04 & 0.36 & 0.15 & 0.11 \\
\multicolumn{1}{c|}{} & mPLUG-Owl2-9.8B & 0.17 & 0.09 & 0.22 & 0.02 & 0.31 & 0.28 & 0.13 \\
\multicolumn{1}{c|}{} & Qwen-VL-Chat-9.6B & 0.28 & 0.20 & 0.28 & 0.14 & 0.33 & 0.30 & 0.11 \\
\multicolumn{1}{c|}{} & Qwen-VL-Plus-[API] & \color[HTML]{0070C0} \textbf{0.46} & 0.51 & {0.38} & {\color[HTML]{0070C0} \textbf{0.53}} & {\color[HTML]{0070C0} \textbf{0.52}} & 0.32 & 0.15 \\
\multicolumn{1}{c|}{\multirow{-5}{*}{LMMs}} & {GPT-4v-[API]} & {0.45} & 0.38 & 0.68 & 0.47 & {0.50} & {\color[HTML]{0070C0} \textbf{0.46}} & {\color[HTML]{0070C0} \textbf{0.23}} \\ \Xhline{1.5\arrayrulewidth}
\multicolumn{1}{c|}{} & ESRA+Qwen-[API] & 0.77 & 0.80 & 0.50 & {\color[HTML]{FF0000} \textbf{0.73}} & {\color[HTML]{FF0000} \textbf{0.68}} & 0.66 & 0.18 \\
\multicolumn{1}{c|}{\multirow{-2}{*}{\begin{tabular}[c]{@{}c@{}}ImageText+\\      LLMs\end{tabular}}} & ESRA+GPT-4-[API] & {\color[HTML]{FF0000} \textbf{0.79}} & 0.76 & 0.62 & 0.72 & 0.63 & {\color[HTML]{FF0000} \textbf{0.79}} & {\color[HTML]{FF0000} \textbf{0.29}} \\ \bottomrule
\end{tabular}
\end{table*}
\subsection{VQA Generation}
We propose a KG-based schema Question-Answering (QA) generator based on the structurally annotation format, as shown in Section~\ref{sec:medical_reasoning_annotation}.

\subsubsection{Medical Knowledge-based Synonym-aware Schema}
Reports from different healthcare information systems often employ various forms of expression, and these differences tend to be further amplified in the Chinese medical terminology. For instance, the presentation of test results in non-laboratory test reports might vary, being described as “impressions”, “pathological diagnosis” or “conclusions”; similarly, reference intervals in laboratory test reports might be referred to as “normal ranges”, “normal intervals” or “standard values”. To reduce the complexity of parsing annotation results, and to increase the diversity of question-asking templates, we have analyzed all the keys from annotations. Together urological experts, we construct a knowledge-based schema that integrates the validated list of synonyms into a comprehensive framework, which includes common vocabulary and their synonyms from different types of reports. Based on this schema, data mapping can be easily performed resulting not only a reduction in ambiguity and errors associated with the interpretation of medical reports due to synonym usage, and an increase in diversity during question generation process to enhanced data interoperability.

\subsubsection{VQA Generator}~\\~
In RJUA-MedDQA dataset, we proposed two tasks: Image Content VQA and Clinical Reasoning VQA.

\textbf{Image Content VQA Generation:}. This task includes entity-based and tabular-based VQA and requires no contextual information. The core of the w/o Content VQA Generation is task generators. Each generator corresponding to multiple sub-task question templates, which differ in their difficulties shown in Table~\ref{tab:qa_generator_example}. Templates contain \texttt{annotated variables} from reports, and \textit{(vocabulary)} in synonym schema. Employing these templates, we generate questions of varying difficulty, which can be enriched through annotations to include complex queries involving dual entities, single-row or multi-row questions. During the formulation of questions, we employ synonym substitution to enhance the generalization ability of our dataset. we have designed templates grounded in our schema and the specific annotation format. 

\textit{Customization:}
In addition to existing QA pairs in the dataset, we provide key-value pair annotations and quadruplets, enabling the construction of customized templates to meet diverse model requirements such as summarization task shown in Table~\ref{tab:qa_generator_example}.

\textbf{Clinical Reasoning VQA Generation:} This task includes multiple choice (MC) and short-answer questions (SA). In each report, diagnosis, status or advice may be explicitly displayed, or could be inferred from image content and corresponding context. For latter one, we match the title from context fact base and use bert-base-chinese \cite{devlin2019bert} embedding and cosine similarity to find the top three similar titles to be distracting options. We have evenly distributed the proportion of each option in the question bank. The answers to the short-answer questions consist of one or more of the correct answers from multiple-choice questions.
% \subsection{Statistics}
% % 从各类图片中，我们等比例采样了数量相同的图片，
% QA pairs /train,test split.
\section{Experiments}
In this section, we report and analyze the extensive experimental results to demonstrate the validity of RJUA-MedDQA.

\subsection{Baseline Models} There are very limited LMM models that have been proposed to effectively solve Chinese medical QA tasks over the images containing both tabular data and textual, where contextual reasoning ability is particularly demanded. We select 5 different multi-modality models and benchmark them on the dataset. The models we have selected cover a broad spectrum of strategies and architectures, effectively illustrating the current state-of-the-art in multimodal understanding i.e. LLaVA-v1.5-7B \cite{liu2023improved}, mPLUG-Owl2-9.8B \cite{ye2023mplugowl2}, Qwen-VL-Chat-9.6B, Qwen-VL-Plus-[API] \cite{bai2023qwenvl} and GPT-4v-[API] \cite{yang2023dawn}. 
In addition to LMMs,  we conduct comparative experiments on a set of strong LLMs by using image-text generated by ESRA method to further investigate the limitations and potential of current LMMs. We will use Qwen-[API] \cite{bai2023qwen} and GPT-4-[API] \cite{openai2023gpt4} for this purpose.
\subsection{Evaluation metrics} The dataset includes diverse tasks which are framed as either short-answer questions or single-choice questions. Our observations indicate that LMMs exhibit preferences for specific answering techniques. For instance, when providing short answers, LMMs tend to rephrase the question before presenting their response. In the case of single-choice questions, the models frequently produce a complete sentence that either corresponds directly with one of the options or is semantically similar to it, sometimes even accompanied by an explanation. Consequently, the reliance on a single evaluation metric or merely comparing the similarity between the model-generated answers and the ground truth (ANLS \cite{biten2019icdar}) can potentially amplify biases in assessment. Therefore, to achieve a more fair evaluation, we have implemented task-specific metrics within the dataset, which allows us to capture the distinct nuances of LMM performance. For Table Numerical Reasoning (NR) QA and Clinical Reasoning Multiple Choice (MC), we adopt soft accuracy which means the predict answer is considered to be correct if it contains the ground truth. For Entity QA, Table QA and Clinical Reasoning Short Answer(SA), we will report both soft accuracy and ROUGE-L \cite{lin-2004-rouge} for comparison. 
\subsection{Results and Analysis}
\begin{figure}[htbp]
\centering\includegraphics[width=1\linewidth]{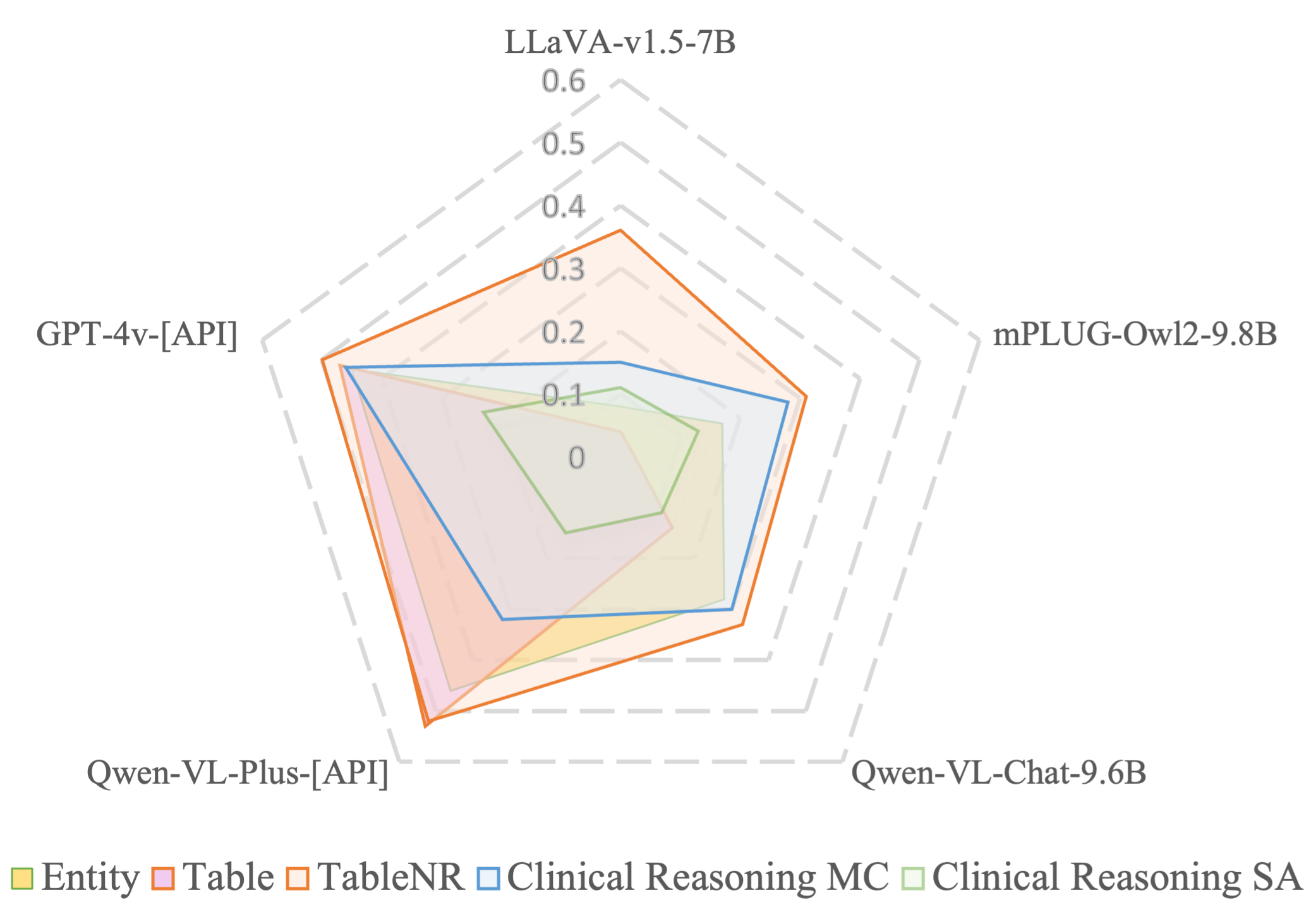}
    \caption{Results of 5 LMMs across 5 tasks}
    \label{fig:9}
\end{figure}
\subsubsection{Overall Results}
We exhaustively evaluate the five models on the existing 5 sub-tasks of RJUA-MedDQA. In Table ~\ref{table:overall}, we present the models' overall performance on various QA tasks namely Entity, Table, NR, Clinical Reasoning MC and SA. Among five LMMs, Qwen-VL-Plus and GPT-4v yield superior results on all tasks, while LLaVA-v1.5-7B and mPLUG-Owl2 demonstrate lower overall performance on Chinese VQA tasks compared to the other models. The performance patterns of Qwen-VL-Plus and GPT-4v on Entity and Table tasks are intriguingly divergent when evaluated using two metrics. One explanation is that Qwen often deviates from the prompt instructions, resulting in longer sentences that negatively impact its RougeL score; while GPT4v tends to change output format such as the decimal place of floats, leading to lower accuracy but higher RougeL score. However, both responses are generally accurate. For entities, where longer text is involved, we prefer to rely on RougeL, while for tables, we opt for Soft Accuracy as a reference. Overall, Qwen-VL-Plus is more proficient in image-content related QAs; while GPT4v shows stronger in-context reasoning ability and perform the best in Clinical Reasoning Tasks. 

Compared to LMMs, ESRA+LLMs yields superior results so far. ESRA+GPT4's performance on most tasks is far better than that of LMMs.  Meanwhile, the performance of all models in Clinical Reasoning SA is not satisfactory.  

\subsubsection{Impact of Image Quality}
For a more in-depth understanding, we provide a comprehensive analysis of the image type (Electronic vs Photo see Table ~\ref{table:level}) and photo quality (High vs Low) on model performance. Although the capabilities of current LMMs are still quite limited and not outperforming traditional image-text with LLMs, an interesting fact emerged when analyzing the impact of image quality on model performance on tabular content (Table~\ref{table:quality}). LMMs more robust to low-quality and diverse-structured images. Table QA requires high image quality since because even minor levels of distortion can lead to a significant drop in model performance. The disparity in GPT-4's ability to process tables is highlighted by a significant performance gap, approximately 0.3, between high and low-quality images, whereas the gap is 0.22 in Qwen. In contrast, such a stark difference in performance is not observed in LMMs. A similar trend is observed when comparing the handling of electronic images and photos. Overall, LMMs often exhibit enhanced performance even with low-quality images, suggesting a level of robustness inherent to multi-modal methodologies that uni-modal OCR-based models may not possess. In Appendix ~\ref{appendix:example}, we present several cases of low-quality images that demonstrate robustness to quality degradation (Figure ~\ref{fig:good}).

\subsubsection{Clinical Reasoning}
Studies indicates that the existence of context definitely improves model performance (Table ~\ref{table:context}). However, the evaluation results reveals that cross-instance understanding and logic reasoning pose a significant challenge for existing LMMs (Table ~\ref{table:reasoning}). GPT-4v exhibits the most advanced reasoning capabilities among all LMMs, yet it falls notably behind ESRA+GPT4 by approximately 0.3. This pattern is mirrored in the comparison between QwenVL+ and QwenVL+ESRA, with an even larger gap of 0.45. It is observed that disease diagnosis tends to be a less complex task than providing clinical advice in table reasoning tasks, as the latter demands multi-step reasoning - a challenge evidently not met by any model except for ESRA+GPT-4. Meanwhile, the SA performance for laboratory and clinical diagnosis by existing models is remarkably poor, with the highest RougeL score being only 0.33. The results indicate that improving the medical report understanding and contexual reasoning capabilities of LLMs can be a significant and promising direction.

\subsection{Case study}
We visualize several loss cases in the experiments of GPT-4v and Qwen-VL-Plus in Appendix ~\ref{appendix:example}. We hope this provide inspirations for future improvements.

\textit{Incorrect Extraction} A large portion of errors are caused by incorrect extraction, an issue that frequently occurs in table QA. For instance, failing to find the correct item name can lead to mistakes when getting reference ranges. A subset of these errors is due to incomplete extraction, where the model correctly identifies the correct entity but fails to include all the necessary information. It's especially hard for models to correctly extract longer text that spans multiple lines (see Figure ~\ref{fig:bad1}). 

\textit{Hallucination} is another common issue, where the model confidently generates text that is not present in the image. We randomly selected 84 questions in Entity QA and 50 questions in Table QA that are unanswerable. Statistics show that models exhibit hallucination issues to varying degrees, with LMMs facing more severe problems. For instance, in Entity Tasks, GPT-4v produced incorrect responses for 54 out of 84 questions; in Table Tasks, the error rate was 35 out of 50 (see Figure ~\ref{fig:bad1}, Figure ~\ref{fig:hallu}).

\textit{Faulty Reasoning} Compared to LLMs, LMMs also need to initially locate the correct content, which adds a certain level of difficulty. (see Figure ~\ref{fig:bad2}, Figure ~\ref{fig:bad4}) In SA tasks, with no hint from options , models are tend to extract text that is directly relevant to the imagery's semantics, rather than fully understanding the test description, conclusions, and corresponding context. For instance, taking the phrase "pre-clinical diagnosis" as the diagnosis for the entire report (see Figure ~\ref{fig:bad3}). Qwen exhibits the same issue, leading to a even inferior performance to GPT-4v.

\begin{table}[]
\caption{Impact of Photo Quality on Model Performance}
\label{table:quality}
\small
\begin{tabular}{c|cc|cc|cc}
\toprule
\multirow{3}{*}{\textbf{Method}} & \multicolumn{2}{c|}{\textbf{Entity}} & \multicolumn{2}{c|}{\textbf{Table}} & \multicolumn{2}{c}{\textbf{TableNR}} \\ \cline{2-7} 
 & High & Low & High & Low & High & Low \\ \cline{2-7} 
 & RougeL & RougeL & Acc & Acc & Acc & Acc \\ \Xhline{1.5\arrayrulewidth}
LLaVA-v1.5 & 0.08 & 0.07 & 0.03 & 0.06 & 0.39 & 0.38 \\
mPLUG-Owl2 & 0.19 & 0.08 & 0.02 & 0.01 & 0.27 & 0.21 \\
Qwen-VL-Chat & 0.33 & 0.29 & 0.08 & 0.09 & 0.33 & 0.26 \\
Qwen-VL-Plus & 0.47 & 0.42 & 0.35 & 0.36 & 0.44 & 0.40 \\
GPT-4v & 0.51 & 0.44 & 0.46 & 0.34 & 0.48 & 0.39 \\ \Xhline{1.5\arrayrulewidth}
ESRA+Qwen & 0.83 & 0.79 & 0.75 & 0.58 & 0.70 & 0.52 \\
ESRA+GPT-4 & 0.82 & 0.77 & 0.71 & 0.40 & 0.61 & 0.48\\ \bottomrule
\end{tabular}
\end{table}

\begin{table}[]
\caption{Performance of Baseline Models on Clinical Reasoning. “D”, “S”, “A” denotes the “Disease”, “Status”, “Advice”}
\label{table:reasoning}
\small
\begin{tabular}{c|cccc|cccc}
\toprule
\multirow{3}{*}{\textbf{Method}} & \multicolumn{4}{c|}{\textbf{Multiple Choice}} & \multicolumn{4}{c}{\textbf{Short Answer}} \\ \cline{2-9} 
 & \multicolumn{2}{c}{Table} & \multicolumn{2}{c|}{Clinical} & \multicolumn{2}{c}{Table} & \multicolumn{2}{c}{Clinical} \\ \cline{2-9} 
 & D & Ad & S & A & D & A & S & A \\ \Xhline{1.5\arrayrulewidth}
LLaVA-v1.5 & 0.1 & 0.13 & 0 & 0.05 & 0.14 & 0.07 & 0.17 & 0.01 \\
mPLUG-Owl2 & 0.18 & 0.17 & 0 & 0.27 & 0.1 & 0.03 & 0.01 & 0.00 \\
Qwen-VL-Chat & 0.24 & 0.21 & 0 & 0.29 & 0.1 & 0 & 0.01 & 0.01 \\
Qwen-VL-Plus & 0.3 & 0.25 & 0.25 & 0.33 & 0.11 & 0.02 & 0.01 & 0.01 \\
GPT-4v & 0.5 & 0.47 & 0.33 & 0.59 & 0.21 & 0.02 & 0.19 & 0.04 \\ \Xhline{1.5\arrayrulewidth}
ESRA+Qwen & 0.75 & 0.65 & 0.33 & 0.52 & 0.06 & 0 & 0.1 & 0.04 \\
ESRA+GPT-4 & 0.86 & 0.87 & 0.5 & 0.67 & 0.22 & 0.05 & 0.33 & 0.06 \\ \bottomrule
\end{tabular}
\end{table}

\section{Conclusion AND FUTURE STUDY }
In this paper, we introduce a new benchmark RJUA-MedDQA to measure
progress on image content extractions and clinical reasoning from visually-rich medical documents in real application. We also proposed a comprehensive data generation pipeline including a image-text restoration method, thoughtful annotation guideline and automatic QA generator which significantly improves efficiency and accuracy. The evaluation results reveals that LMMs
demonstrate robustness to quality poor-quality images but still struggle with complex reasoning tasks involving cross-instance understanding and strong logical link. Our research aims to help improve multi-modal understanding of medical documents and aid in healthcare applications. In future study, We will also focus on the three areas of opportunity discovered in this paper, and explore approaches that can improves contextual reasoning tasks.

\bibliographystyle{ACM-Reference-Format}
\bibliography{sample-base}

\appendix
\section{Statistics of Image}
\label{diease:statistics}
\begin{figure*}
    \centering
    \includegraphics[width=0.9\linewidth]{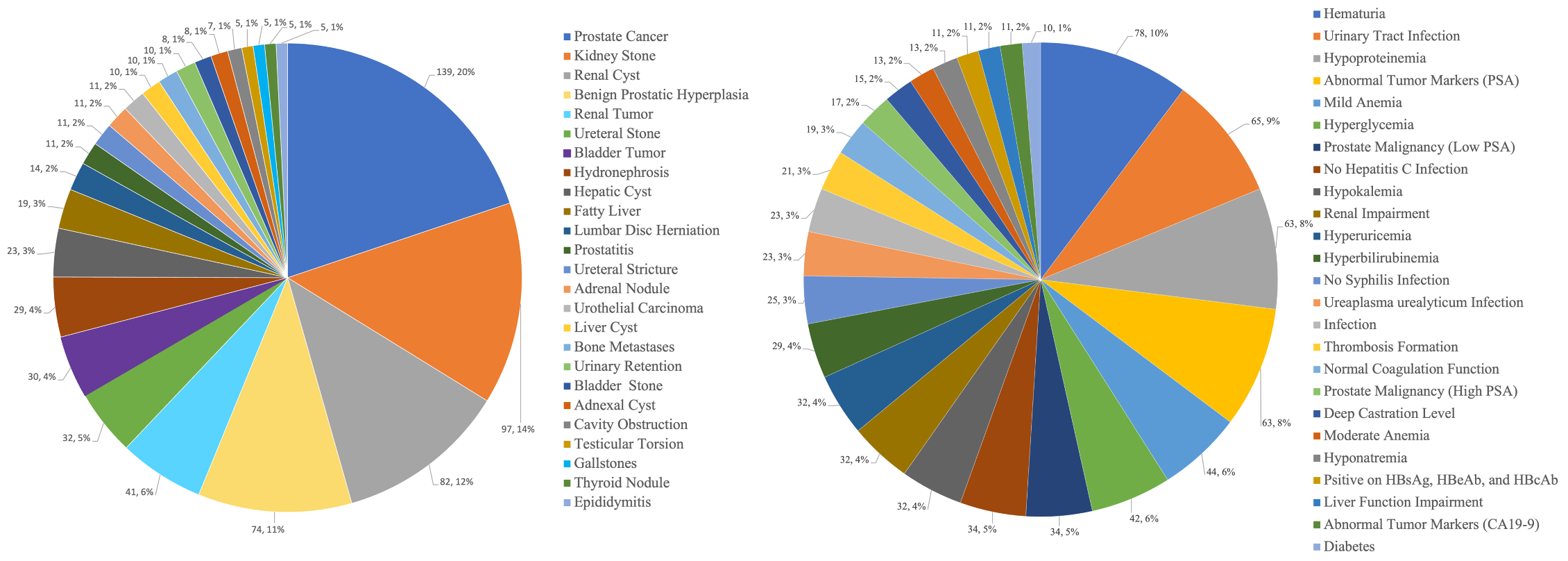}
    \caption{Top 25 Diseases in Laboratory Reports (Left) and Diagnostic Reports (Right)}
    \label{fig:5}
\end{figure*}

\begin{figure*}
    \centering
    \includegraphics[width=1\linewidth]{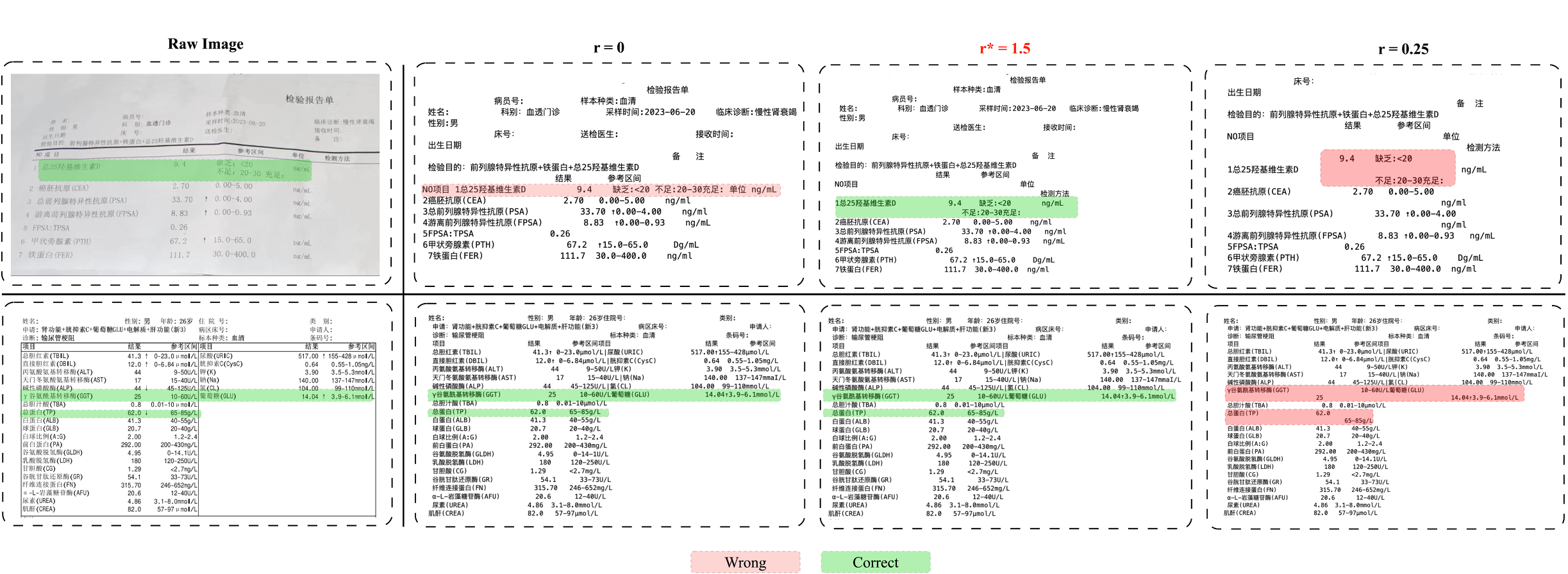}
    \caption{The Impact of Different \textit{r} on the output structure of ESRA Structure: An Illustration}
    \label{fig:ESRA}
\end{figure*}

The RJUA-MedDQA dataset comprises 2000 images, among which 402 are screenshots, 619 are scanned PDFs, and the remaining 979 are photos taken by patients in real-world medical consultations. Within the photos category, we classified each image based on its quality. The evaluation of low-quality images is dependent on two sensors: Completeness Sensor and Angle Perception Sensor. 
\begin{itemize}[leftmargin=4.5mm]
    \item [1.] Completeness Sensor identifies the paper's four corners in an image. If less than three corners are detected, the image is deemed incomplete.
    \item [2.] Angle Perception Sensor captures skewness. Given Four points of the image $[x_{0,0},x_{0,1},y_{1,0},y_{1,1}]$, if the angle between $\text{line}_{x_{0,0},x_{0,1}}$ and $\text{line}_{x_{1,0},x_{1,1}}$ is greater than 15 degrees, it is deemed skewness.
\end{itemize}
A photo that meets any either of the criteria is considered to be of low quality. Overall, there are 386 low-quality images, shown in Table ~\ref{table:image_type}

RJUA-MedDQA covers approximately 166 unique diseases and diagnoses for laboratory report, 334 for diagnostic report (non-laboratory report). Of these, 80\% are urological-realted diseases, and we have listed the top 25 diseases for both report types, shown in Figure ~\ref{fig:5}.

\section{Data Processing}
\subsection{Implement Details of ESRA Method}
\label{hyperparameter}
As mentioned in Section 4.1, $r^*$ is
a discount coefficient ranging from 0 to 1. By setting this hyper-parameter $r^*$, the adhesion of text segments between lines can be mitigated, thereby increasing readability. We have tested various settings for $r$ ranging from 0 to 1 (see Figure ~\ref{fig:ESRA}). It can be observed that for electronic raw images (PDF), the impact of $r$ is negligible, whereas for photographs, $r$ results in a more distinct segmentation of each line. We have determined that $r^* = 0.15$ is the optimal value for processing both electronic and photographed images.

$l$ is an expansion coefficient from 0 to 1. By setting the hyper-parameter $l$, we control the coefficient that determines the number of spaces per line in the overall layout output. $l^*=0.7$ is optimal.
\begin{table}[h]
\caption{Statistics of Medical Documents in RJUA-MedDQA}
\label{table:image_type}
\small
\begin{tabular}{cc|c|c}
\toprule
\multicolumn{1}{c|}{\textbf{Image Type}} & \textbf{Quality} & \textbf{\# Images} & \textbf{\# kv pairs} \\ \Xhline{1.5\arrayrulewidth}
\multicolumn{1}{c|}{\multirow{2}{*}{Photo}} & High & 593 & \multirow{2}{*}{7603} \\
\multicolumn{1}{l|}{} & Low & 386 &  \\ \hline
\multicolumn{1}{c|}{PDF} & High & 619 & 6895 \\ \hline
\multicolumn{1}{c|}{Screenshot} & High  & 402 & 2351 \\ \Xhline{1.5\arrayrulewidth}
\multicolumn{2}{c|}{\textbf{Overall}} & 2000 & 16849 \\ \bottomrule
\end{tabular}
\end{table}
\subsection{Statistics of RJUA-MedDQA}
\label{context}
In this section, we present an overview of the fundamental statistics for the RJUA-MedDQA. This includes the total count of questions and images, the average length of questions, the average length of answers (see Table ~\ref{table:statisticsQA})

\begin{table}[h]
\small
\caption{Statistics of QAs in RJUA-MedDQA}
\label{table:statisticsQA}
\begin{tabular}{cc|c|c}
\toprule
\multicolumn{1}{c|}{\textbf{Task}} & \textbf{\begin{tabular}[c]{@{}c@{}}Answer \\ Type\end{tabular}} & \textbf{\# Questions} & \textbf{\begin{tabular}[c]{@{}c@{}}Averge\\  Answer Length\end{tabular}} \\ \Xhline{1.5\arrayrulewidth}
\multicolumn{1}{l|}{\multirow{2}{*}{Entity}} & Single & 16887 & 20.8 \\
\multicolumn{1}{l|}{} & Multi & 8705 & 50.2 \\ \hline
\multicolumn{1}{l|}{\multirow{2}{*}{Table}} & Single & 21784 & 5.4 \\
\multicolumn{1}{l|}{} & Multi & 10892 & 11.9 \\ \hline
\multicolumn{1}{l|}{\multirow{2}{*}{TableNR}} & Single & 10892 & 2.2 \\
\multicolumn{1}{l|}{} & Multi & 930 & 17.3 \\ \hline
\multicolumn{1}{l|}{\multirow{2}{*}{Reason}} & MC & 1228 & 12.1 \\
\multicolumn{1}{l|}{} & SA & 846 & 17.2 \\ \Xhline{1.5\arrayrulewidth}
\multicolumn{2}{c|}{\textbf{Overall}} & 72164 & 17.1 \\ \bottomrule
\end{tabular}
\end{table}

We have listed the types and quantities of annotated contexts corresponding to laboratory reports and clinical reports, 178 contexts in total. 
Notice that laboratory reports do not include contexts for disease treatments since because it is difficult to extract accurate treatment plans from a single laboratory report. Meanwhile, the lengths of contexts for diagnostic reports are longer than those for laboratory reports. The reason is that the conclusions of laboratory reports are usually related to specific indicators and involve shorter deductive chains, as illustrated in Table ~\ref{contextexample}. In contrast, diagnostic reports include more detailed content based on descriptive observations, such as the dimensions and conditions of the tissues observed, contributing to a variety of therapeutic approaches.

Overall, the clinical reasoning task requires that models have the capability to integrate both the content of images and specific medical information in given contexts in order to derive the correct statement.

\begin{table}[h] 
\caption{Statistics of Annotated Contexts. "Exam" denotes Examination}
\label{contextexample}
\small
\begin{tabular}{cc|c|c}
\toprule
\multicolumn{1}{c|}{\textbf{\begin{tabular}[c]{@{}c@{}}Report \\ Type\end{tabular}}} & \textbf{\begin{tabular}[c]{@{}c@{}}Context \\ Type\end{tabular}} & \textbf{\# Contexts} & \textbf{\begin{tabular}[c]{@{}c@{}}Averge \\ Context Length\end{tabular}} \\ \Xhline{1.5\arrayrulewidth}
\multicolumn{1}{c|}{\multirow{4}{*}{Table}} & Exam-Disease & 55 & 63.1 \\
\multicolumn{1}{c|}{} & Exam-Status & 7 & 86.3 \\
\multicolumn{1}{c|}{} & Disease-Status  & 1 & 86 \\
\multicolumn{1}{c|}{} & Disease-Advice & 46 & 71.5 \\ \hline
\multicolumn{1}{c|}{\multirow{5}{*}{Clinical}} & Exam & 7 & 61 \\
\multicolumn{1}{c|}{} & Exam-Disease & 3 & 29 \\
\multicolumn{1}{c|}{} & Disease-Exam & 16 & 100.8 \\
\multicolumn{1}{c|}{} & Disease-Status & 3 & 408 \\
\multicolumn{1}{c|}{} & Disease-Treatment & 40 & 310.5 \\ \Xhline{1.5\arrayrulewidth}
\multicolumn{2}{c|}{\textbf{Overall}} & 178 & 135.1 \\ \bottomrule
\end{tabular}
\end{table}

\section{Experiments}
\subsection{Impact of Question Difficulty}
Based on the data presented in Table ~\ref{table:level}, it is apparent that for Large Language Models (LLMs), the span of the text containing the answer still influences model performance. The scores for Multi-span are lower than those for Single-span across all tasks, particularly for Table and Table NR task. However, LMMs seem to be less affected, especially for questions within the same row of a table structure, where LMMs (e.g., gpt-4v and qwenvlplus) appear to perform better. Nevertheless, when it comes to reasoning capabilities, there is a significant difference between single indicator anomaly detection and multi-indicator anomaly detection. The performance of all models on multi-indicator anomaly detection tasks is not ideal. GPT models tend to perform slightly better compared to QWEN models in these tasks.
\begin{table}[h]
\small
\caption{Impact of Question Difficulty on Model Performance. 'S' denotes 'Single'; 'M' denotes 'Multi'}
\label{table:level}
\begin{tabular}{c|cc|cc|cc}
\toprule
 &
  \multicolumn{2}{c|}{\textbf{Entity}} &
  \multicolumn{2}{c|}{\textbf{Table}} &
  \multicolumn{2}{c}{\textbf{TableNR}} \\ \cline{2-7} 
 &
  \multicolumn{1}{c}{\textbf{S}} &
  \multicolumn{1}{c|}{\textbf{M}} &
  \multicolumn{1}{c}{\textbf{S}} &
  \multicolumn{1}{c|}{\textbf{M}} &
  \textbf{S} &
  \textbf{M} \\ \cline{2-7} 
\multirow{-3}{*}{\textbf{Method}} &
  RougeL &
  RougeL &
  Acc &
  Acc &
  Acc &
  Acc \\ \Xhline{1.5\arrayrulewidth}
LLaVA-v1.5-7B &
  0.08 &
  0.07 &
  0.03 &
  0.06 &
  0.37 &
  0.04 \\
mPLUG-Owl2 &
  0.18 &
  0.16 &
  0.01 &
  0.03 &
  0.36 &
  0.20 \\
Qwen-VL-Chat &
  0.31 &
  0.26 &
  0.14 &
  0.16 &
  0.33 &
  0.22 \\
Qwen-VL-Plus &
  0.41 &
  {\color[HTML]{4472C4} \textbf{0.54}} &
  {\color[HTML]{4472C4} \textbf{0.53}} &
  {\color[HTML]{4472C4} \textbf{0.56}} &
  {\color[HTML]{4472C4} \textbf{0.54}} &
  0.19 \\
GPT-4v &
  {\color[HTML]{4472C4} \textbf{0.44}} &
  0.47 &
  0.47 &
  0.5 &
  0.50 &
  {\color[HTML]{4472C4} \textbf{0.28}} \\ \Xhline{1.5\arrayrulewidth}
ESRA+Qwen &
  0.84 &
  0.65 &
  0.75 &
  {\color[HTML]{FF0000} \textbf{0.72}} &
  {\color[HTML]{FF0000} \textbf{0.70}} &
  0.38 \\
ESRA+GPT-4 &
  {\color[HTML]{FF0000} 0.85} &
  {\color[HTML]{FF0000} \textbf{0.76}} &
  0.75 &
  0.71 &
  0.64 &
  {\color[HTML]{FF0000} \textbf{0.55}} \\ \bottomrule
\end{tabular}
\end{table}

\subsection{Impact of Image Type}
\label{appendix:imagetype}
\begin{table}[h]
\small
\caption{Impact of Image Type on Model Performance. "E" denotes "Electronic(PDS/Screenshot)"; "P" denotes "Photo"}
\label{table:imagetype}
\begin{tabular}{c|cc|cc|cc}
\toprule
\multirow{3}{*}{\textbf{Method}} & \multicolumn{2}{c|}{\textbf{Entity}} & \multicolumn{2}{c|}{\textbf{Table}} & \multicolumn{2}{c}{\textbf{TableNR}} \\ \cline{2-7} 
 & E & P & E & P & E & P \\ \cline{2-7} 
 & RougeL & RougeL & Acc & Acc & Acc & Acc \\ \hline
LLaVA-v1.5-7B & 0.09 & 0.06 & 0.04 & 0.05 & 0.38 & 0.32 \\
mPLUG-Owl2 & 0.20 & 0.10 & 0.02 & 0.02 & 0.34 & 0.25 \\
Qwen-VL-Chat & 0.31 & 0.23 & 0.17 & 0.09 & 0.34 & 0.28 \\
Qwen-VL-Plus & 0.47 & \color[HTML]{4472C4} \textbf{0.44} & \color[HTML]{4472C4} \textbf{0.63} & 0.36 & \color[HTML]{4472C4} \textbf{0.54} & \color[HTML]{4472C4} \textbf{0.42} \\
GPT-4v & \color[HTML]{4472C4} \textbf{0.52} & 0.41 & 0.52 & \color[HTML]{4472C4} \textbf{0.40} & 0.51 & 0.41 \\ \Xhline{1.5\arrayrulewidth}
ESRA+Qwen & 0.76 & 0.81 & \color[HTML]{FF0000} \textbf{0.83} & 0.55 & \color[HTML]{FF0000} \textbf{0.72} & 0.52 \\
ESRA+GPT-4 & \color[HTML]{FF0000} \textbf{0.78} & \color[HTML]{FF0000} \textbf{0.83} & 0.82 & \color[HTML]{FF0000} \textbf{0.56} & 0.64 & \color[HTML]{FF0000} \textbf{0.54}\\ \bottomrule
\end{tabular}
\end{table}

\subsection{Statistics on Hallucination}
\begin{figure}[b]
    \centering
    \includegraphics[width=0.9\linewidth]{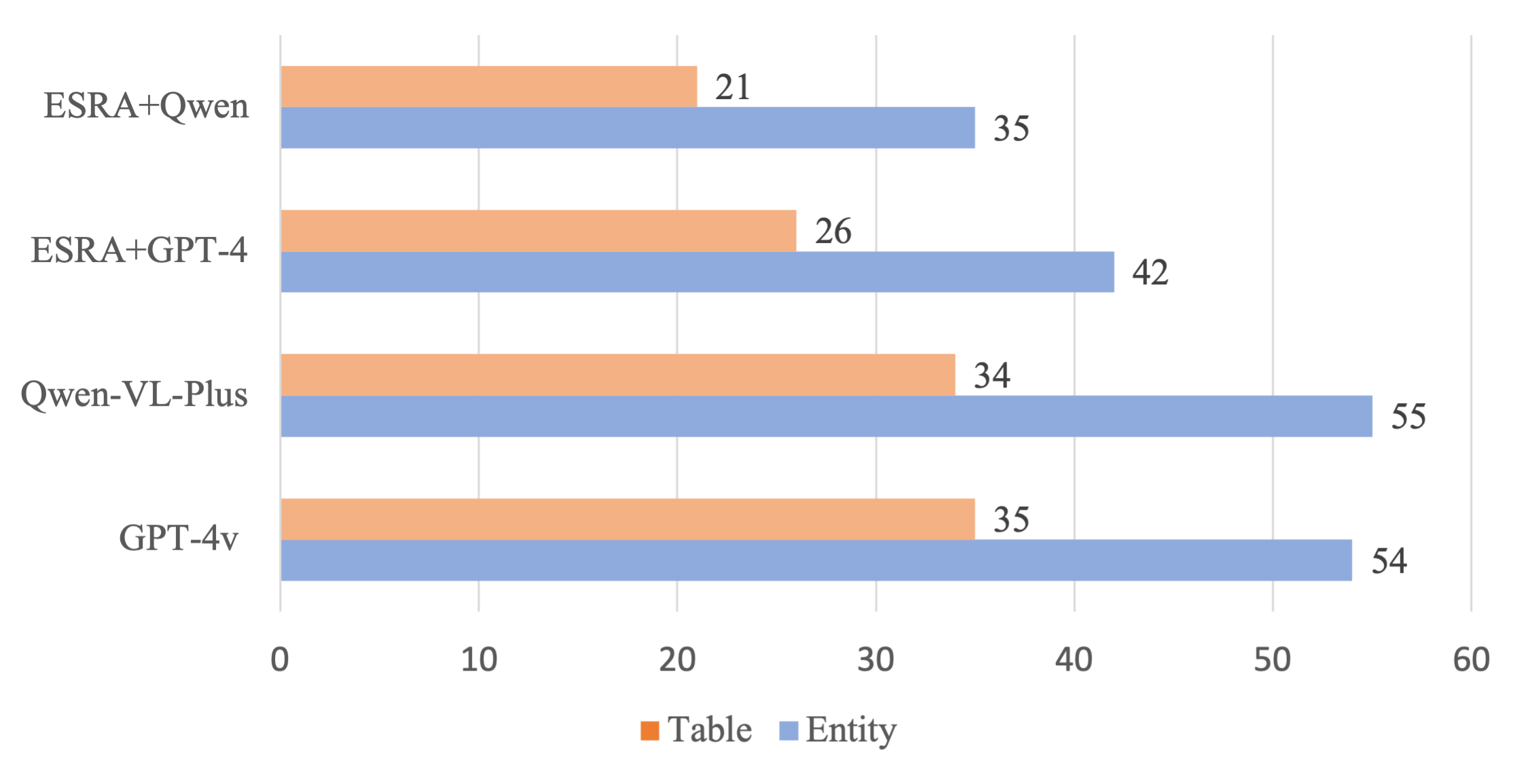}
    \caption{Statistics on Errors in Hallucination}
    \label{fig:hallu}
\end{figure}

\begin{table*}[h]
\small
\caption{Impact of Contexts on Model Performance}
\label{table:context}
\begin{tabular}{c|cccccc|cccccc}
\toprule
\multirow{3}{*}{Method} & \multicolumn{6}{c|}{Multiple Choice} & \multicolumn{6}{c}{Short Answer} \\ \cline{2-13} 
 & \multicolumn{3}{c|}{Table} & \multicolumn{3}{c|}{Clinical} & \multicolumn{3}{c|}{Table} & \multicolumn{3}{c}{Clinical} \\ \cline{2-13} 
 & Q & C+Q & \multicolumn{1}{c|}{$\Delta$} & Q & C+Q & $\Delta$ & Q & C+Q & \multicolumn{1}{c|}{$\Delta$} & Q & C+Q & $\Delta$ \\  \Xhline{1.5\arrayrulewidth}
Qwen-VL-Plus & 0.29 & 0.3 & \multicolumn{1}{c|}{+0.01} & 0.28 & 0.31 & +0.03 & 0.08 & 0.09 & \multicolumn{1}{c|}{+0.01} & 0.01 & 0.01 & +0.00 \\ 
GPT-4v & 0.40 & 0.48 & \multicolumn{1}{c|}{+0.08} & 0.38 & 0.45 & +0.07 & 0.13 & 0.21 & \multicolumn{1}{c|}{+0.08} & 0.1 & 0.19 & +0.09 \\ \Xhline{1.5\arrayrulewidth}
ESRA+Qwen & 0.65 & 0.72 & \multicolumn{1}{c|}{+0.07} & 0.45 & 0.48 & +0.03 & 0.04 & 0.05 & \multicolumn{1}{c|}{+0.01} & 0.06 & 0.01 & +0.02 \\ 
ESRA+GPT-4 & 0.70 & 0.86 & \multicolumn{1}{c|}{+0.16} & 0.62 & 0.66 & +0.04 & 0.14 & 0.22 & \multicolumn{1}{c|}{+0.08} & 0.22 & 0.30 & +0.08 \\ \bottomrule
\end{tabular}
\end{table*}

\label{appendix:example}
\begin{figure*}
    \centering
    \includegraphics[max width=\linewidth]{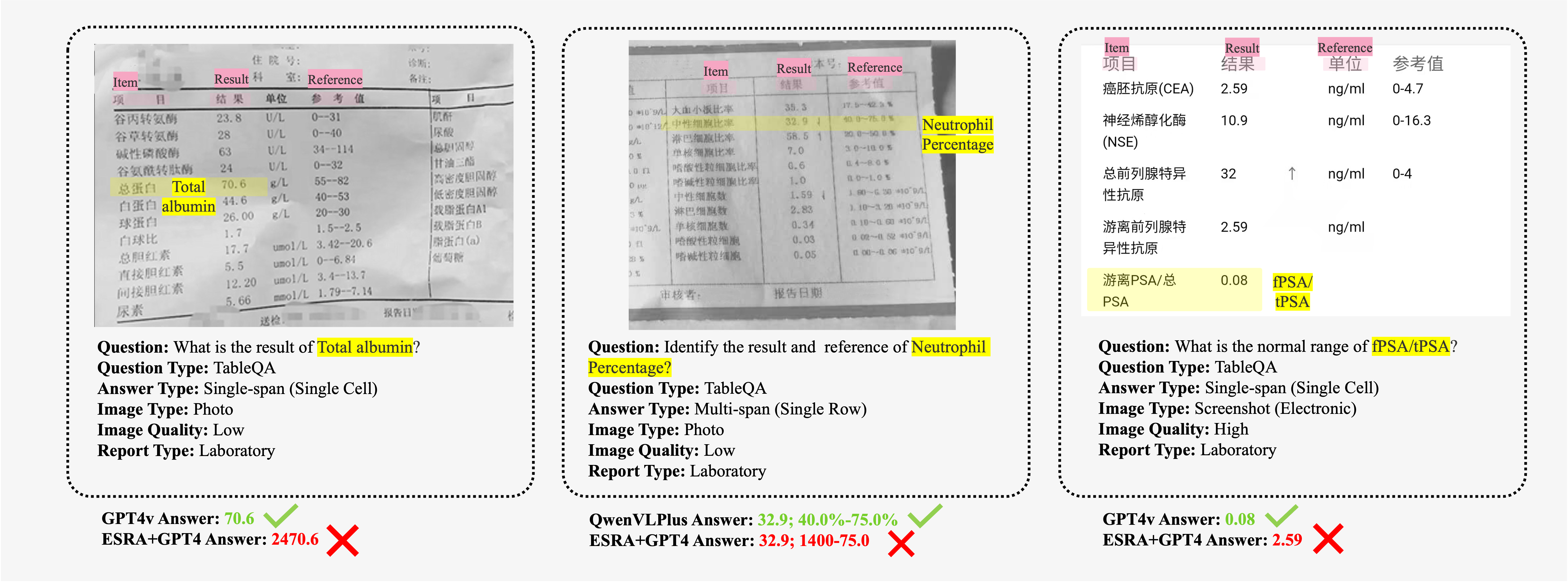}
    \caption{Good cases for low quality image}
    \label{fig:good}
\end{figure*}

\begin{figure*}
    \centering
    \includegraphics[max width=\linewidth]{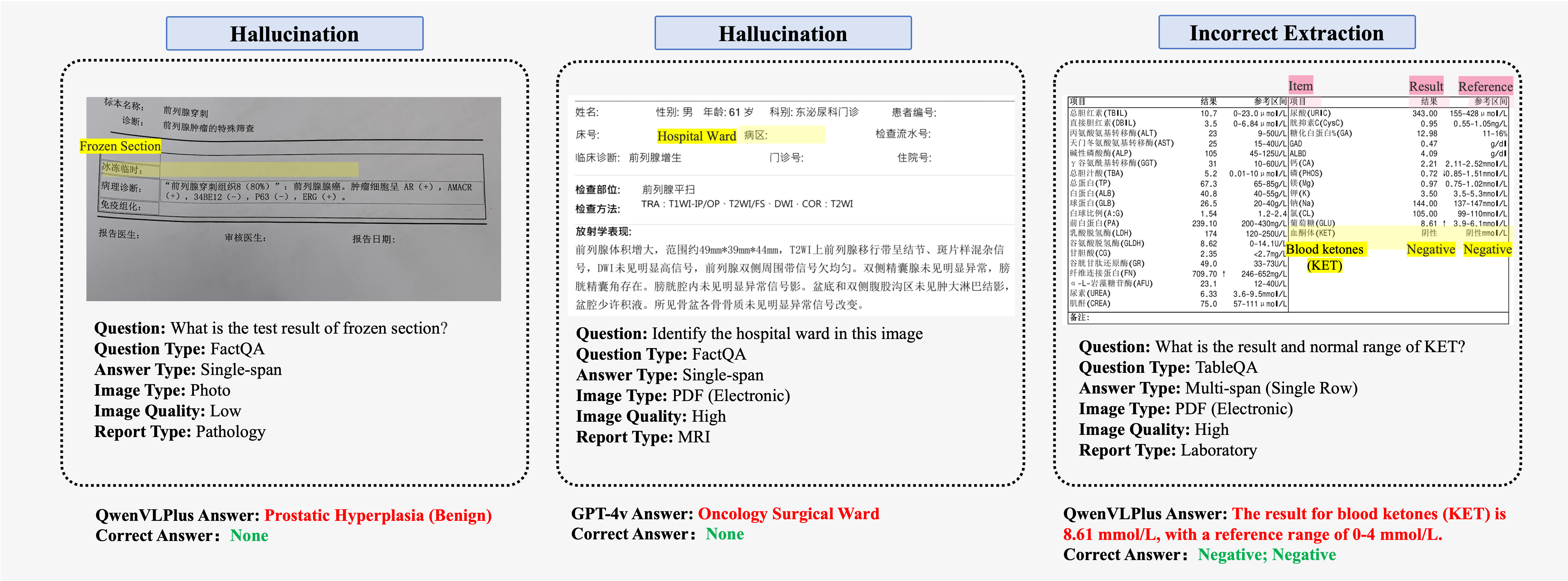}
    \caption{Bad cases for Image Content QA}
    \label{fig:bad1}
\end{figure*}

\begin{figure*}
    \centering
    \includegraphics[max width=\linewidth]{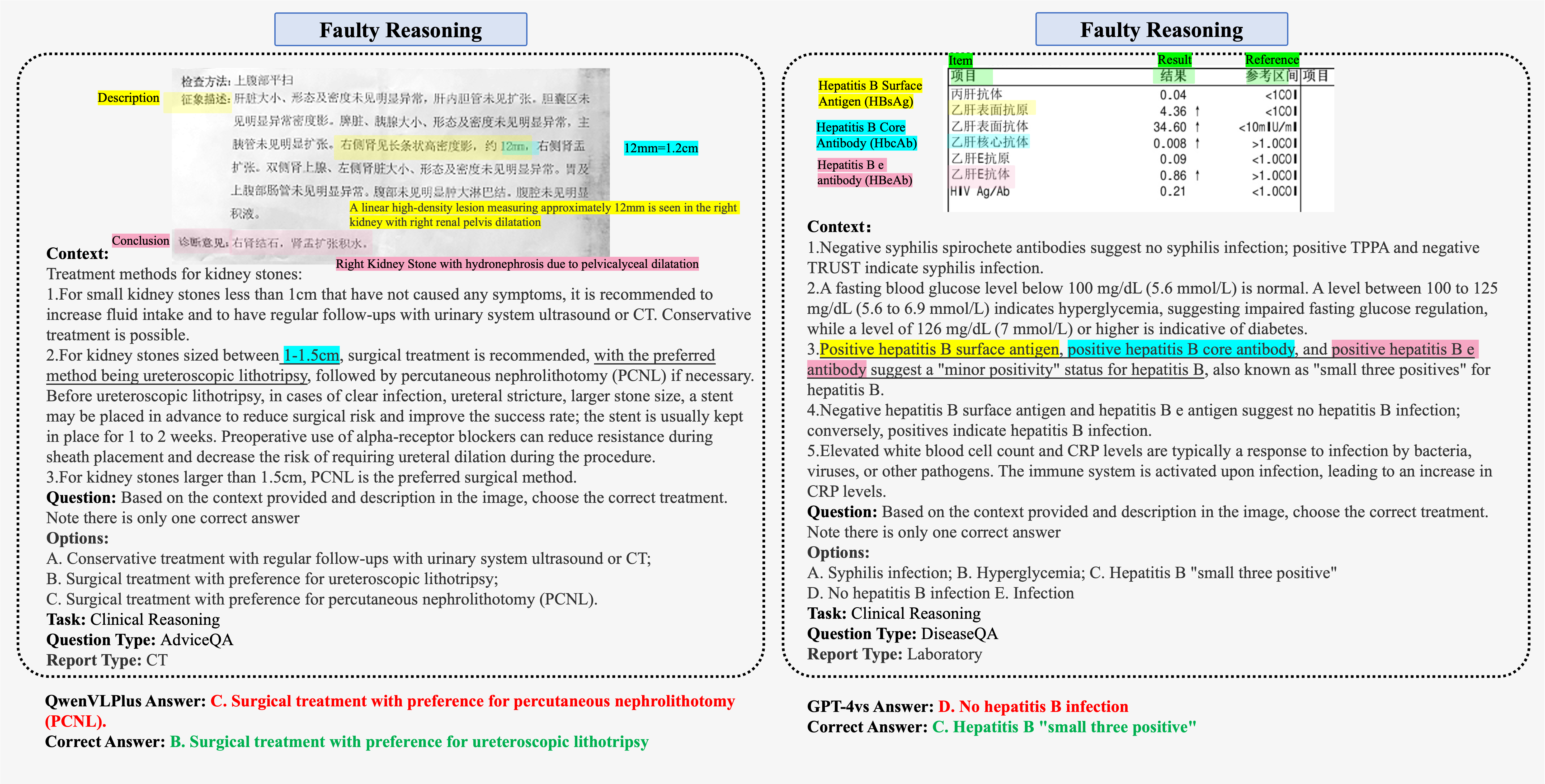}
    \caption{Bad cases for Medical Reasoning MC}
    \label{fig:bad2}
\end{figure*}

\begin{figure*}
    \centering
    \includegraphics[max width=\linewidth]{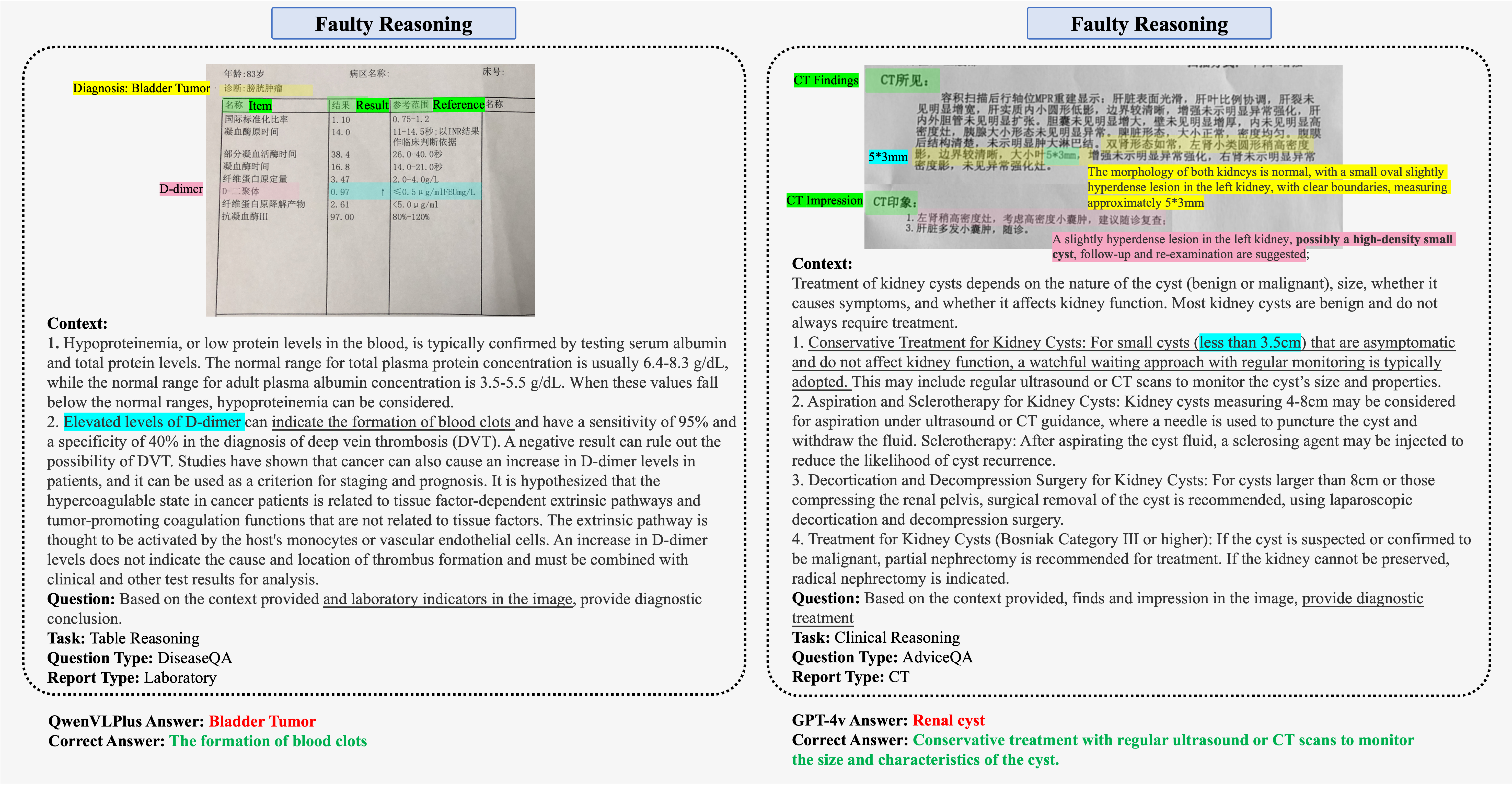}
    \caption{Bad cases for Medical Reasoning SA}
    \label{fig:bad3}
\end{figure*}

\begin{figure*}
    \centering
    \includegraphics[max width=\linewidth]{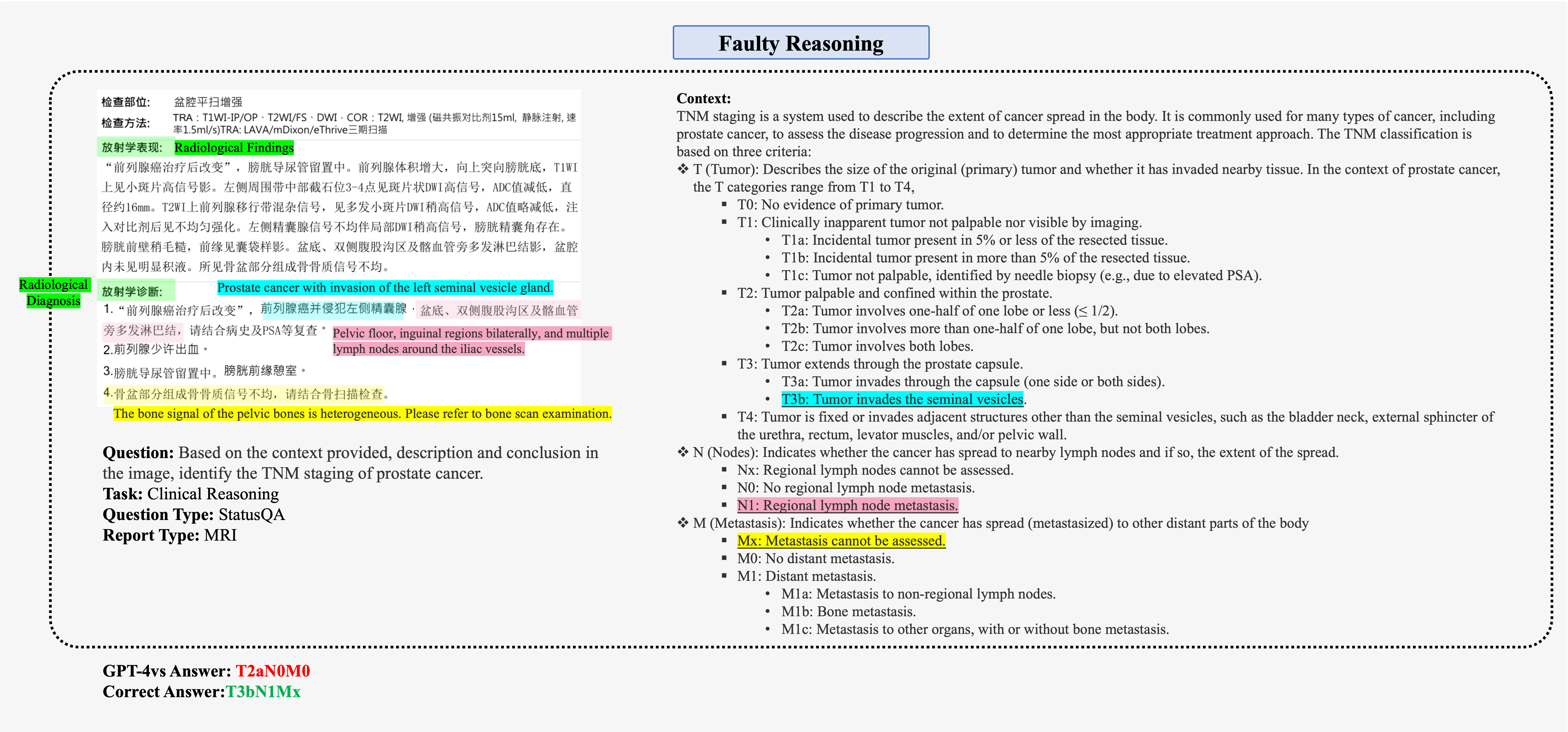}
    \caption{Bad cases for Medical Reasoning SA}
    \label{fig:bad4}
\end{figure*}

\begin{table*}
\footnotesize
\caption{Examples of Annotated Contexts}
\begin{tabular}{l|l|l|l}
\toprule
\textbf{Type} &
  \textbf{Title} &
  \textbf{Type} &
  \textbf{Description} \\  \Xhline{1.5\arrayrulewidth}
Lab &
  Mild Anemia &
  Exam-Disease &
  \begin{tabular}[c]{@{}l@{}}A hemoglobin level of 90g/L to 120g/L indicates the presence of mild anemia; \\ A hemoglobin level of 60g/L to   90g/L indicates the presence of moderate anemia.\end{tabular} \\ \hline
Lab &
  \begin{tabular}[c]{@{}l@{}}Malignant \\ Prostate Tumor \\ (Low PSA)\end{tabular} &
  Exam-Disease &
  \begin{tabular}[c]{@{}l@{}}In patients with malignant prostate tumors, high levels of PSA are associated with a higher \\tumor burden; low levels of PSA indicate a better treatment response and a lower tumor burden.\end{tabular} \\ \hline
Lab &
  \begin{tabular}[c]{@{}l@{}}Poor Prognosis \\ for Testicular Tumors\end{tabular} &
  Disease-Status &
  \begin{tabular}[c]{@{}l@{}}Elevated LDH levels suggest a   poor prognosis for patients with testicular tumors; \\ pre-chemotherapy LDH levels can be used as a prognostic factor for overall survival (OS) in the  \\ intermediate-risk group of non-seminomas. In seminomas, lactate dehydrogenase   (LDH) has \\ been proven to be an additional adverse prognostic factor.\end{tabular} \\ \hline
Lab &
  \begin{tabular}[c]{@{}l@{}}Thrombosis Formation \\ (High Coagulation Risk)\end{tabular} &
  Disease-Advice &
  \begin{tabular}[c]{@{}l@{}}Thrombosis formation can be   treated with low-molecular-weight heparin or further monitored \\ using thromboelastography or TAT/PIC levels.\end{tabular} \\ \hline
Clinical &
  Renal Cysts Treatment &
  Disease-Treatment &
  \begin{tabular}[c]{@{}l@{}}
  Treatment of kidney cysts   depends on the nature of the cyst (benign or malignant), size, \\ 
  whether it causes symptoms, and whether it affects kidney function. Most kidney cysts are \\ 
  benign and do not always require treatment.\\ 
  1. Conservative Treatment: For small cysts (less than 3.5cm) that are asymptomatic and do not \\ 
  affect kidney function, a watchful waiting approach with regular monitoring is typically \\
  adopted. This may include regular ultrasound or CT to monitor the cyst’s size and properties.\\
  2. Aspiration and Sclerotherapy: Kidney cysts measuring 4-8cm are considered for aspiration \\ 
  under ultrasound or CT guidance, where a needle is used to puncture the cyst and withdraw \\the
  fluid. Sclerotherapy: After aspirating the cyst fluid, a sclerosing agent may be injected \\to 
  reduce the likelihood of cyst recurrence.\\
  3. Decortication and Decompression Surgery: For cysts larger than 8cm or those compressing\\
  the renal pelvis, surgical removal  of   the cyst is recommended, using laparoscopic \\decortication and decompression surgery.\\ 
  4. Treatment for Kidney Cysts (Bosniak Category III or higher): If the cyst   is suspected or \\
  confirmed to be malignant, partial nephrectomy is recommended   for treatment. If \\
  the kidney cannot be preserved, radical nephrectomy is   indicated.\end{tabular} \\ \bottomrule
\end{tabular}
\end{table*}

\begin{table*}[]
\small
\caption{Examples of Medical Knowledge-based Synonym-aware Schema}
\begin{tabular}{l|l|l}
\toprule
\multicolumn{1}{c|}{\textbf{Topic}} & \multicolumn{1}{c|}{\textbf{Key Entity}} & \multicolumn{1}{c}{\textbf{Synonyms}} \\ \Xhline{1.5\arrayrulewidth}
 & Name & Patient \\ \cline{2-3} 
 & Age & DateofBirth \\ \cline{2-3} 
 & Date & \begin{tabular}[c]{@{}l@{}}Examination Date,  Test Date,    Report Date,  Writing Date,   \\      Date Received,  Audit Date,  Submission Time,  Specimen Accptance Time, \\      Report,  Audit\end{tabular} \\ \cline{2-3} 
\multirow{-4}{*}{\textbf{Basic Information}} & pre-test Diagnosis & Clinical Diagnosis \\ \hline
 & Item & Examination Item, Test Item,    Indicator,  Laboratory Indicator \\ \cline{2-3} 
 & Result & Examination Result, Laboratory Result,  Test Result \\ \cline{2-3} 
\multirow{-3}{*}{\textbf{Laboratorary}} & Range & Referene Range, Normal Range, Standard Range, Normal Interval \\ \hline
 & Description & CT Appearance, \\ \cline{2-3} 
\multirow{-2}{*}{\textbf{CT}} & Conclusion & CT Diagnosis,  CT   Impression \\ \hline
{\color[HTML]{333333} \textbf{Endoscopic}} & Description & {\color[HTML]{333333} Cystoscopy Description,  Endoscopic Description} \\ \hline
\textbf{Ultrasound} & Description & {\color[HTML]{333333} Ultrasound Description,  Ultrasound Findings,  Ultrasound Contrast, TRUS} \\ \hline
 & Description & {\color[HTML]{333333} \begin{tabular}[c]{@{}l@{}}Frozen Section,  Immunohistochemistry, Macroscopic   Examination, \\      Microscopic Findings\end{tabular}} \\ \cline{2-3} 
\multirow{-2}{*}{\textbf{Pathology}} & Conclusion & {\color[HTML]{333333} \begin{tabular}[c]{@{}l@{}}Pathological Diagnosis, Supplementary   Pathological Diagnosis,  \\      Comprehensive Evaluation\end{tabular}} \\ \hline
 & Description & \begin{tabular}[c]{@{}l@{}}Radiological Appearance,  Imaging Appearance, Observation   records,\\      Examination findings, Inspection findings, Film indication, Impression note\end{tabular} \\ \cline{2-3} 
\multirow{-2}{*}{\textbf{General}} & Conclusion & \begin{tabular}[c]{@{}l@{}}Radiological Diagnosis,  Imaging Diagnosis,  Imaging Conclusion,\\ Clinical diagnosis, Examination diagnosis,  Imaging findings, \\ Conclusions and Recommendations,  Diagnosis,  Diagnostic Opinion, Impression\end{tabular} \\ \bottomrule
\end{tabular}
\end{table*}
\end{document}